\documentclass[10pt,journal,compsoc]{IEEEtran}
\ifCLASSINFOpdf
\else
\fi
\usepackage{booktabs}
\usepackage{multirow}
\usepackage{rotfloat}
\usepackage{pdflscape} 
\usepackage{longtable} 
\usepackage{caption}
\usepackage{amssymb}
\usepackage{comment}
\usepackage{amsmath}
\usepackage{tabularray}
\usepackage{xspace} %
\usepackage{soul}
\usepackage[thicklines]{cancel}
\usepackage[pagebackref=true,breaklinks=true,colorlinks,citecolor=blue,urlcolor=blue,linkcolor=blue,bookmarks=false]{hyperref}
\usepackage{graphicx}  
\usepackage{float}  
\usepackage{subfig}
\usepackage[normalem]{ulem}
\usepackage[numbers]{natbib}
\usepackage{algorithm}
\usepackage{algpseudocodex}
\usepackage{tcolorbox}
\usepackage[thicklines]{cancel}
\hypersetup{hidelinks,
	colorlinks=true,
	allcolors=black,
	pdfstartview=Fit,
	breaklinks=true}
\usepackage{graphicx}  
\usepackage{float}  
\usepackage{subfig}

\makeatletter
\DeclareRobustCommand\onedot{\futurelet\@let@token\@onedot}
\def\@onedot{\ifx\@let@token.\else.\null\fi\xspace}
\def\eg{\emph{e.g}\onedot} 
\def\ie{\emph{i.e}\onedot}

\def\etc{\emph{etc}\onedot}

\def\etal{\emph{et al}\onedot}
\def\versus{\emph{vs}\onedot}
\makeatother

\newtcolorbox[auto counter]{promptbox}[1][]{
  colback=orange!5!white,
  colframe=orange!75!black,
  fonttitle=\bfseries,
  title=Prompt~\thetcbcounter: #1
}

\begin{document}
\title{Global-Local Monte Carlo Tree Search in Vision-Language Models for Text-to-3D Indoor Scene Generation}

\author{
        Mengshi Qi,~\IEEEmembership{Member,~IEEE},
        Wei Deng,
        Xianlin Zhang,
        Huadong Ma,~\IEEEmembership{Fellow,~IEEE}
\thanks{This work is partly supported by the Funds for the NSFC Project under Grant 62572072, Beijing Natural Science Foundation (L243027). (\emph{Corresponding author: Mengshi Qi~(email:~qms@bupt.edu.cn)})}
\thanks{M. Qi, W. Deng, X. Zhang and H. Ma are with the State Key Laboratory of Networking and Switching Technology, Beijing University of Posts and Telecommunications, China.}
}

\markboth{}%
{TITLE}

\IEEEtitleabstractindextext{
\begin{abstract}
Large Vision-Language Models (LVLMs) have achieved significant reasoning performance in various tasks.
However, there are few studies on text-to-3D indoor scene generation with LVLMs. The main challenge is that prevailing LVLM-based methods employ chain-of-thought~(CoT) sequential decision mechanisms that cannot revise earlier decisions, causing error propagation.
In this paper, we consider the task as a planning problem constrained by spatial and layout commonsense.
To solve this problem, we model it as a tree search problem with global and local trees, which differs from existing sequential decision-making approaches.
In the global tree, we place each object iteratively and explore multiple attempts like humans furnishing a room, where the problem space is represented as a tree.
To effectively search the tree, we propose a hierarchical scene representation and a progress reward model (PRM)-guided Monte Carlo Tree Search (MCTS) method.
The hierarchical representation abstracts a scene into room level, region level, floor object level, and supported object level.
The PRM-guided MCTS method uses the PRM to prune unnecessary branches and the MCTS algorithm to balance exploration and exploitation to get an optimal solution with fewer attempts.
In the local tree, it further decomposes the placement of each object into finer sub-steps, including the specific placement parameters.
To make the whole appearance of the scene consistent, we leverage pre-trained diffusion image generative models to predict textures for all the objects in the scene.
As existing benchmarks for text-to-3D indoor scene generation remain limited in scale and diversity, we collect a new large-scale diverse dataset that contains 65 scene types and 3,250 instructions with diverse sizes, layouts, and styles, named \emph{3DTindo-bench}, to better assess the capability of the state-of-the-art models. Our experimental results show that our method generates more realistic 3D scenes than state-of-the-art approaches, and our method surpasses the best baseline by approximately 14\% on average performance scores. Our source code and dataset are open-sourced at \href{https://github.com/dw-dengwei/TreeSearchGen}{https://github.com/dw-dengwei/TreeSearchGen}.

\end{abstract}

}
\maketitle
\IEEEdisplaynontitleabstractindextext
\IEEEpeerreviewmaketitle
\section{Introduction}
\label{sec:intro}

3D indoor scene generation~\cite{AnyHome,ATISS,CommonScenes,DiffuScene,Graph-to-3d,InstructScene,LayoutGPT,PhyScene,SceneHGN,SceneFormer,HoloDeck} involves creating realistic 3D environments automatically. In recent years, several studies~\cite{AnyHome,DiffuScene,InstructScene,LayoutGPT,HoloDeck} have investigated natural language-driven scene creation via the generative models, \ie, text-to-3D scene generation, as textual descriptions provide intuitive user interfaces. Thus, synthesizing high-quality 3D indoor scenes holds significant value across multiple applications, including interior design, 3D gaming, virtual/augmented reality, and embodied AI.

\begin{figure}
    \centering
    \includegraphics[width=0.8\linewidth]{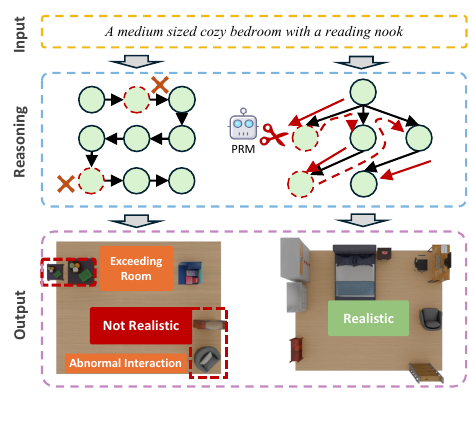}
    \vspace{-7mm}
    \caption{
    Illustration of the text-to-3D indoor scene synthesis task. 
    (Left) Previous methods rely on chain-structured reasoning, which accumulates errors.
    (Right) Our proposed PRM method regards this task as a tree-search problem, which can backtrack to fix errors. 
    }
    \vspace{-5mm}
    \label{fig:intro-fig}
\end{figure}

Achieving high-fidelity 3D indoor scene synthesis fundamentally requires capturing reasonable and physically-consistent spatial relationships among scene objects to describe the 3D layout.
Conventional data-driven methodologies learn joint probability distributions of indoor environments and their corresponding textual descriptions from 3D datasets~\cite{DiffuScene,InstructScene}.
Nevertheless, acquiring comprehensive 3D scene datasets remains resource-intensive, yielding constrained dataset sizes that limit model generalization and robustness.
Contemporary investigations increasingly leverage Large Vision-Language Models (LVLMs)~\cite{AnyHome,LayoutGPT,HoloDeck,GALA3D} to tackle text-to-3D scene generation.
LVLMs, pre-trained on massive cross-modal datasets, demonstrate remarkable capacity for interpreting complex textual directives and synthesizing 3D arrangements by leveraging learned indoor furnishing knowledge patterns.
For example, LayoutGPT~\cite{LayoutGPT} outputs comprehensive layout parameters containing size, position, and orientation specifications for individual objects.
Despite these advancements, as illustrated in Figure~\ref{fig:intro-fig} (Left), prevailing LVLM architectures employ chain-of-thought sequential decision mechanisms that proceed left-to-right during inference~\cite{ToT}, rendering them ill-suited for 3D spatial layout reasoning.
Specifically, such auto-regressive paradigms cannot revise earlier decisions.
Consequently, misplacement of a preceding object propagates errors throughout subsequent generation stages.
Thus, advancing LVLM capabilities for robust 3D scene generation continues to present significant challenges.

It is worth noting that effective 3D scene synthesis necessitates tree-structured search strategies rather than linear sequential approaches.
When humans furnish indoor environments, we typically position furniture items progressively while considering multiple viable placement options per item.
Humans select one candidate for the current item and proceed to position subsequent objects, iteratively continuing until all components are placed.
If an object occupies an unsuitable location, it obstructs placement of subsequent items, prompting reconsideration of prior decisions.
Therefore, as shown in Figure~\ref{fig:intro-fig} (Right), the solution space naturally forms a tree structure.
Each tree layer represents a distinct object, while nodes within the same layer denote alternative placement configurations.
The path spanning from root to terminal layer constitutes a complete layout solution.
Our goal involves searching this tree structure to identify optimal configurations that satisfy spatial constraints, placement heuristics, non-overlapping requirements, and non-floating specifications.
Nevertheless, exhaustive traversal of expansive tree structures introduces prohibitive computational complexity, particularly for scenes containing numerous objects.
Consequently, we identify a critical need for search algorithms that can efficiently explore the solution space while maintaining quality. Furthermore, alongside rapid advancements in text-to-3D indoor scene synthesis, establishing comprehensive evaluation benchmarks has become increasingly critical.
To our knowledge, existing evaluation protocols, such as the LayoutVLM-Bench~\cite{LayoutVLM} covering only 11 scene types with 33 instructions and lacking size and style variations, employ substantially limited sample sizes when assessing different approaches, as indicated in Table~\ref{tab:bench-basic}.

To address these limitations, we introduce a novel Progress Reward Model (PRM)-guided Monte Carlo Tree Search (MCTS) framework to advance LVLM capabilities for 3D indoor scene synthesis.
Scenes containing numerous elements present substantial challenges for deep tree traversal.
Our central insight recognizes that scene architectures naturally lend themselves to hierarchical decomposition.
To this end, we decompose a scene into room-level, region-level, floor-object-level, and supported-object-level hierarchies, forming a hierarchical scene representation.
This representation functions as a semantic intermediary bridging textual directives and 3D scene outputs, enabling region-wise object placement that significantly reduces computational overhead.
Subsequently, we integrate PRM with MCTS algorithm to generate layouts within each region.
The PRM is adopted to evaluate intermediate states and prune unpromising branches, thereby avoiding exhaustive search.
The MCTS algorithm incorporates selection, expansion, simulation, and backpropagation steps to effectively balance exploration and exploitation during tree search.
Additionally, we leverage pre-trained diffusion-based image synthesis frameworks to generate object textures that ensure unified visual appearance across the complete scene.

Furthermore, we propose 3DTindo-bench, the large-scale diverse benchmark for text-to-3D indoor scene generation.
In contrast to existing benchmarks such as LayoutVLM-Bench~\cite{LayoutVLM} (11 scene types, 33 instructions, no size or style coverage), ours covers 65 scene categories with 3,250 textual instructions, spanning diverse room sizes, inter-object relations, and stylistic specifications.
Beyond the dataset, we introduce an LVLM-powered evaluation framework that assesses generated scenes across five dimensions: physical reasonability, semantic reasonability, layout reasonability, instruction alignment, and aesthetic consistency.

More importantly, this work is an extension of our CVPR conference paper~\cite{Global-local}.
In contrast to the previous version, we improved the tree search method from Depth First Search~(DFS) to PRM-guided MCTS.
This enhancement avoids blindly and exhaustively searching the problem tree.
Moreover, the original paper does not consider appearance consistency.
This paper generates texture for all the objects in a scene to achieve appearance consistency.
Additionally, we collect a new large-scale benchmark~(3DTindo-bench), which involves more scene types and data samples compared with the original version. We also evaluate the improvements in enhancing DFS in the PRM-guided MCTS algorithm and the effectiveness of texture generation. 

In summary, our principal contributions encompass three key aspects:

\par\textbf{(1)} We propose to solve the 3D indoor scene generation with LVLMs as a tree search problem, and introduce a hierarchical scene representation methodology that bridges textual instructions with 3D scene outputs while simultaneously diminishing computational requirements;

\par\textbf{(2)} We introduce a novel PRM-guided MCTS framework for text-to-3D scene generation. The PRM prunes unpromising branches to reduce computational overhead, while the MCTS algorithm balances exploration and exploitation during searching the solution space;

\par\textbf{(3)} We establish a new 3DTindo-bench, the large-scale diverse benchmark for text-to-3D indoor scene generation, and empirically demonstrate that our approach surpasses state-of-the-art techniques by approximately 14\% on average performance scores.

\section{Related Work}
\label{sec:related}

\noindent\textbf{Data-driven 3D Scene Generation.}
3D scene generation is not a trivial task that can be solved by directly applying a 3D object generation model to produce multiple objects.
The primary difficulty is layout generation.
It needs to model spatial relationships between objects.
Traditional methods train generative models (\eg, GANs~\cite{Giraffe, BlockGAN}, VAEs~\cite{Graph-to-3d, CommonScenes, SceneHGN}, diffusion models~\cite{CommonScenes, DiffuScene, InstructScene, PhyScene}, and autoregressive models~\cite{ATISS, SceneFormer}) on 3D scene datasets~\cite{3D-FRONT, 3D-FUTURE, 3DSSG}.
For instance, CommonScenes~\cite{CommonScenes} employs a variational auto-encoder to predict the layout, and a latent diffusion model to generate object shapes. Huang~\etal~\cite{MIDI} propose a multi-instance attention mechanism in the diffusion model to capture inter-object relations. Meng~\etal~\cite{LT3SD} propose to employ a latent tree representation to hierarchically encode scene geometry and details, capturing both coarse structures and fine details. Feng~\etal propose CASAGPT~\cite{CasaGPT}, which uses cuboid primitives to represent and arrange 3D objects. Previous methods are data-driven, requiring training on limited-scale 3D scene datasets (\eg, 3D-FRONT with 18k scenes), which constrains generalization to unseen scenes. In contrast, our method is knowledge-driven, leveraging the commonsense and spatial reasoning embedded in pretrained LVLMs to generate layouts without training on annotated scene data.

\noindent\textbf{Text-to-3D Scene Generation.}
3D scene datasets are considerably small compared with 3D object datasets (\eg, 3D-FRONT~\cite{3D-FRONT} 18k+ \versus Objaverse~\cite{objaverse} 800k+ and Objaverse-XL~\cite{Objaverse-XL} 10M+).
Thus, models trained on small-scale datasets are less robust and limited to a few types of scenes.
Recently, several works~\cite{LayoutVLM,LayoutGPT,HoloDeck,AnyHome} utilize the common sense within LVLMs to perceive user-provided instructions and generate layouts.
They also use the CLIP~\cite{CLIP} to retrieve the most relevant 3D assets from a 3D object database.
For instance, LayoutGPT~\cite{LayoutGPT} directly outputs the scene layout formatted as Cascading Style Sheets (CSS), including size, location, and orientation for each object.
Zhang~\etal propose Scene Language~\cite{The-Scene-Language}, including a program defining the hierarchical and relational organization of scene entities, natural language words describing the semantic category of each entity, and embeddings encoding their visual characteristics.
Sun~\etal propose LayoutVLM~\cite{LayoutVLM} which leverages the semantic understanding of LVLMs and introduces a scene layout representation that supports differentiable optimization to ensure physically feasible arrangements.
Zhou~\etal propose the GALA3D~\cite{GALA3D} for compositional text-to-3D scene generation that enables precise and user-friendly control.
GALA3D leverages an LLM to create an initial scene layout and employs a layout-guided 3D Gaussian representation to generate 3D content.
HoloDeck~\cite{HoloDeck} and AnyHome~\cite{AnyHome} propose to generate scene graphs, which represent the objects as nodes and spatial relationships as edges, and propose rule-based algorithms to convert the scene graph into the room layout.
Bai~\etal propose FreeScene~\cite{FreeScene} with a Vision-Language Model-based Graph Designer, which converts them into a graph representation. Previous LVLM-based methods generate layouts in chain-like reasoning process, without exploring multiple placement options or recovering from suboptimal intermediate decisions. In contrast, we formulate layout generation as a tree search problem in this work, enabling multi-path exploration and backtracking over placement choices.

\noindent\textbf{Reasoning in LLMs/LVLMs}. Slow-thinking in LLMs enhances reasoning during test time, which is very important in problem-solving, decision-making, and critical thinking~\cite{reason-survey}.
LLMs and LVLMs can be prompted with a standard input-output paradigm~\cite{GPT3, GPT4, PaLM,LLaMA,Qwen-VL,LLaVA}.
However, language models are trained to generate coherent language sequences.
Such a simple prompting method falls short when the task is complex and requires multi-step reasoning.
To enable step-by-step reasoning, Wei~\etal~\cite{CoT} propose Chain-of-Thought (CoT), which enforces the language models to output intermediate thoughts.
Wang~\etal~\cite{self-consistency} propose a novel strategy called self-consistency to enhance CoT prompting.
This method samples a diverse set of reasoning trajectories and selects the most consistent answer by marginalizing over all possible paths.
CoT-based methods significantly improve the performance on reasoning tasks.
Nevertheless, other tasks require exploring multiple alternatives at each intermediate step rather than just picking one.
To this end, Yao~\etal~\cite{ToT} propose Tree-of-Thoughts (ToT), which maintains a tree of thoughts to find solutions for a complex task.
To avoid an exponential search space, the beam algorithm~\cite{beam_search} is applied, by maintaining the top $N$ highest-probability thoughts at each reasoning step.
It can be regarded as a pruned breadth-first search algorithm, which discards low-probability paths~\cite{tts-survey}.
However, previous tree-based reasoning methods rely on exhaustive search with DFS/BFS or standard MCTS without intermediate state evaluation.
The Monte Carlo Tree Search (MCTS) algorithm~\cite{mcts} relies on stochastic sampling to assess reasoning steps, guiding the algorithm towards the paths with better outcomes.
Thus, we propose PRM-guided MCTS, which integrates a Progress Reward Model to evaluate intermediate layout states, enabling efficient search with early pruning of low-quality branches.

\section{Preliminary}\label{sec:formulation}

\noindent\textbf{Problem Formulation.}
The goal of text-to-3D indoor scene generation is to place 3D assets, which are retrieved from a supporting 3D object database, following user instructions and physical constraints. Formally, given a textual instruction $x$, a supporting 3D object dataset $\mathcal{D}=\{d_1, d_2, \cdots, d_M\}$, and a retrieval function $f$ that yields an asset in the supporting dataset $d_i\in\mathcal{D}$ according to the asset's attributes, \eg, description, category, \etc.
We aim to use an LVLM to output a scene $S=\{o_1, o_2, \cdots, o_N |o_i=(c_i, s_i, p_i, r_i, d_i) \}$ with $N$ objects, where each object $o_i$ is represented with its category $c_i$, size $s_i$, position $p_i$, orientation $r_i$ and a 3D asset $d_i$ retrieved from $\mathcal{D}$.
We evaluate the output scene $S$ using a scoring function that quantifies how well $S$ matches the input instruction $x$ and commonsense.

\noindent\textbf{Monte Carlo Tree Search (MCTS).}
MCTS~\cite{mcts} is a searching algorithm that efficiently explores large decision spaces by incrementally constructing a search tree.
It has been widely adopted in decision making~\cite{mcts}.
MCTS builds a tree where each node represents a state and each edge represents an action that transitions from one state to another.
The algorithm iteratively grows and evaluates the tree through the following four steps:

\noindent\textbf{(1) Selection.}
Starting from the root node, the algorithm recursively selects a child node according to a tree policy until reaching a leaf node.
The most widely used tree policy is the Upper Confidence bounds for Trees (UCT)~\cite{UCT}:
\begin{equation}
C_{\textrm{select}} = \mathop{\arg\max}_{C'\in \textrm{children}(C)} \overline{v}_{C'} + \epsilon\sqrt{\frac{\ln n_C}{n_{C'}}},
\label{eq:standard-uct}
\end{equation}
where $\overline{v}_{C'}=v_{C'}/n_{C'}$ is the average value of node $C'$, $v_{C'}$ is the accumulated value, $n_{C'}$ is the visit count of $C'$, $n_C$ is the visit count of the parent node $C$, and $\epsilon$ controls the exploration--exploitation trade-off.
The first term $\overline{v}_{C'}$ encourages exploitation of high-value nodes, while the second term $\epsilon\sqrt{\ln n_C / n_{C'}}$ promotes exploration of less-visited nodes.

\noindent\textbf{(2) Expansion.}
Once a leaf node is selected, child nodes are generated by a policy model $\pi$ to expand the tree:
\begin{equation}
\textrm{children}(C) = \{C_1', C_2', \dots, C_k' \mid C_i'\sim\pi(C)\},
\end{equation}
where $k$ is the number of children and $C_i'$ is a child node sampled from $\pi$ given the current state $C$.

\noindent\textbf{(3) Simulation.}
From a newly expanded node, the algorithm performs a stochastic simulation (also called a rollout) by sampling actions until reaching a terminal state $T$.
The terminal state is evaluated by a scoring function to obtain a value:
\begin{equation}
v = \textrm{score}(T),
\end{equation}
where $v$ reflects the quality of the terminal state $T$.

\noindent\textbf{(4) Backpropagation.}
The value $v$ is propagated back along the path from the selected node to the root, updating the accumulated value and visit count of each node:
\begin{equation}
\begin{aligned}
v_i &\gets v_i + v,\\
n_i &\gets n_i + 1,
\end{aligned}
\end{equation}
where $v_i$, $n_i$ are the accumulated value and visit count of the $i$-th node on the path.

After a predefined computation budget, MCTS returns the trajectory leading to the node with the highest average value, providing an approximate solution to the search problem.

\noindent\textbf{Layout Tree.} We model layout generation as a tree search problem. As described in Section~\ref{sec:intro}, sequentially placing objects with multiple options at each step naturally forms a tree structure.
Formally, we define a \textbf{layout tree} as:
\begin{equation}
\begin{gathered}
\mathcal{T}=(\mathcal{V},\mathcal{E}),\qquad
\mathcal{V}=\bigcup_{l=0}^{N}\mathcal{V}_l,\\
\mathcal{E}=\{v_{l-1}\to v_l\mid v_{l-1}\in\mathcal{V}_{l-1},\,v_l\in\mathcal{V}_l\}.
\end{gathered}
\end{equation}
where $\mathcal{V}_l$ is the set of nodes at depth $l$, each node $v_l\in\mathcal{V}_l$ represents a layout with $l$ objects placed, and each edge $v_{l-1}\to v_l$ represents adding one object to the layout. The root $\mathcal{V}_0=\{v_0\}$ is the empty layout, and each path $v_0\to v_1\to\cdots\to v_N$ corresponds to a complete layout of $N$ objects.
To manage the search complexity, we decompose the layout tree into two levels: a \textbf{global layout tree} $\mathcal{T}_{\mathrm{global}}$ for object-level placement, and a \textbf{local layout tree} $\mathcal{T}_{\mathrm{local}}$ for the sub-steps of placing each individual object.
Both are described in detail in Section~\ref{sec:tree}.

\begin{figure*}
    \centering
    \includegraphics[width=0.9\linewidth]{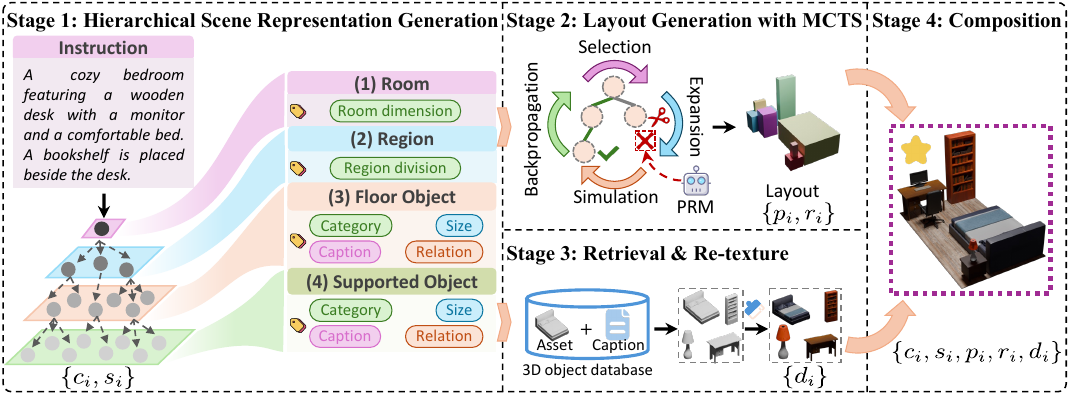}
    \vspace{-2mm}
    \caption{Overview of our method. We first generate a hierarchical scene representation from a textual instruction. Then, we conduct layout generation with the MCTS-based reasoning method and object retrieval \& re-texture for consistent texture. Finally, we compose the objects via the generated layout, which defines where to put the objects and how to orient them.}
    \vspace{-2mm}
    \label{fig:overview}
\end{figure*}

\noindent\textbf{Overview.} 
As illustrated in Figure~\ref{fig:overview}, in stage 1 (Sec.~\ref{sec:hierarchical}), our method generates a hierarchical scene representation, which serves as a proxy $P$ to link the instruction $x$ and the 3D scene $S$: $p^{x\to S}=p^{P\to S}(S|P)p^{x\to P}(P|x)$.
The hierarchical scene representation abstracts the scene structure.
It defines what objects should be placed into the scene and how we arrange the objects.
It decomposes a scene into many functional areas, and we independently process the areas and combine them to form the whole scene.
In this stage, the category $c_i$ and size $s_i$ of objects are generated.
In stage 2 (Sec.~\ref{sec:tree} and Sec.~\ref{sec:MCTS}), we search on the layout tree defined above with the PRM-guided MCTS algorithm to generate the layout for the scene given the proxy $P$.
In this stage, the position $p_i$ and orientation $r_i$ are determined.
In stage 3 (Sec.~\ref{sec:texture}), we retrieve assets from the given supporting dataset $\mathcal{D}$ and generate texture for all the objects that follow the instruction $x$.
In this stage, the 3D asset $d_i$ for each object is determined.
Finally, we compose the objects via the generated layout to get the generated scene $S$.
All the parameters for each object $o_i\in S$ are respectively generated in different stages.

\section{Hierarchical Scene Representation}
\label{sec:hierarchical}

Due to the large gap between the textual instruction and the 3D scene, we utilize an LVLM to produce a proxy that bridges them.
The proxy is a hierarchical structure that is derived from the nature of indoor scenes. Specifically, as shown in Figure~\ref{fig:overview}, the hierarchy includes room level, region level, floor object level, and supported object level.
The LVLM starts from the user's instruction and generates the levels step by step.
During the process, the LVLM is prompted to follow the instruction and consider the commonsense.

\noindent\textbf{(1) Room level.}
The first level stands for the whole room.
The room is parameterized by its length and width $(l,w)$.

\noindent\textbf{(2) Region level.}
The second level is the region level.
In this level, we divide a scene into multiple functional regions.
Each region is parameterized by its region type and dimensions $(t_i, l_i,w_i)$.
The region type $t_i$ is consistent with the kind of room, ~\eg, \texttt{rest region} for \texttt{bedroom}.

\noindent\textbf{(3) Floor object level.}
The floor object level represents the objects that are typically placed on the floor.
We independently generate floor objects for each region.
A region contains a set of objects $S'=\{o_1',o_2',\cdots,o_N'\}$ and spatial relationships between an object and an anchor $E'=\{e'_{ia}|o_i',o_a'\in S'\}$.
We first prompt the LVLM to generate the category $c_i$, size $s_i$, and style description $x_i$ for each floor object $o_i$.
The style description $x_i$ contains the style or texture requirements for the object.
All the categories and style descriptions are coherent with each other.
We then prompt the LVLM to determine an anchor and spatial relationships.
The anchor $o_a'\in S'$ is the object that represents the main semantics of this region, \eg, \texttt{bed} for \texttt{rest region} in \texttt{bedroom}.
The anchor $o_a'$ and the relation set $E'$ help us to determine the layout for the non-anchor objects in the region.

\noindent\textbf{(4) Supported object level.}
The supported objects are typically placed on a semantically-consistent floor object, \eg, \texttt{alarm clock} on \texttt{nightstand}.
Similar to the floor object level, we also instruct the LVLM to generate a set of objects and spatial relationships between the supported anchor object and others.

\section{Layout tree for PRM-guided MCTS}\label{sec:tree}

\begin{figure}
\centering
\includegraphics[width=1.\linewidth]{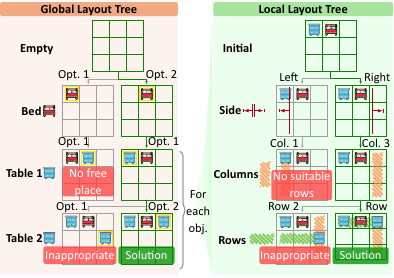}
\vspace{-6mm}
\caption{
Structure of the global layout tree (left) and local layout tree (right).
(Left) We define the global layout tree starting from an empty node that does not contain any objects.
We sequentially place the objects, for instance bed, in table 1 and table 2, until we find a solution to place all objects.
The placement for each individual object could have multiple options, hence the problem space is a tree structure.
(Right) For the placement of each individual object.
We define the local tree, which decomposes the process of determining the placement parameters into three steps, including determining the side (left/right), columns (1/2/3), and rows (1/2/3) for placing the new object.
}
\vspace{-3mm}
\label{fig:global-local}
\end{figure}

Before introducing the PRM-guided MCTS algorithm, we first define the searching space for the algorithm.
As defined in Section~\ref{sec:formulation}, the layout tree models the layout generation process as a tree structure, where each node represents a partial layout and each edge represents placing a new object.
To solve the layout generation problem, we perform the PRM-guided MCTS algorithm on the global and local layout trees.
Specifically, as shown in Figure~\ref{fig:global-local}, the global layout tree manages the generation process globally, whose nodes represent an individual object.
The local layout tree decomposes the process into smaller sub-tasks, whose nodes represent the specific placement parameters for an object. In the following sections, we first introduce the definition of the global layout tree and later the local layout tree.

\subsection{Global Tree}
\noindent\textbf{Layout Node Definition.}
We formulate the global tree nodes as the placement set of objects:
\begin{equation}
C_l=\{(p_1,r_1), (p_2,r_2), \cdots, (p_l,r_l)\},
\end{equation}
where $l$ is the depth of the node in the tree, $p_i$ and $r_i$ are the position and orientation for the $i^{\textrm{th}}$ object, respectively.
$C_l$ represents the layout for the first $l$ objects.
The global tree starts from a root node $C_0$ in layer $0$, which represents an empty layout.
For each node $C_l$ in the depth of $l$, it has multiple child nodes, as there are many choices to place the $l+1^{\mathrm{th}}$ node into the existing layout of $C_l$.
For generating the layout for a region, we aim to search for a trajectory from the root node $C_0$ to a terminal node $C_{N'}$, where $N'$ is the number of objects in a region.
The searching space is consequently a tree structure.

\noindent\textbf{Layout Edge Definition.}
In the global tree, the layout nodes are connected with directed edges, from the parent node to its children:
\begin{equation}
C_{l-1}\to C_{l}=\pi_{\mathrm{local}}(C_{l-1}, P),
\end{equation}
where $\pi_{\mathrm{local}}$ is the local tree solver, which generates the placement parameters $p_l$ and $r_l$, \ie, the child node $C_l$, for the $l^{\textrm{th}}$ object based on the existing layout of $C_{l-1}$ and the proxy $P$.

\subsection{Local Tree}
In this stage, we define a local tree to model the generation process $p_l,r_l\sim \pi_{\textrm{local}}(C_l,P)$ that inserts a new object into an existing layout.
To generate orientation $r_l$ for the object, we leverage the knowledge in the LVLM to determine the orientation from the following options: \texttt{face to anchor}, \texttt{back to anchor}, \texttt{face the same direction as the anchor}, and \texttt{face the opposite direction of the anchor}.

\begin{figure}
    \includegraphics[width=1.\linewidth]{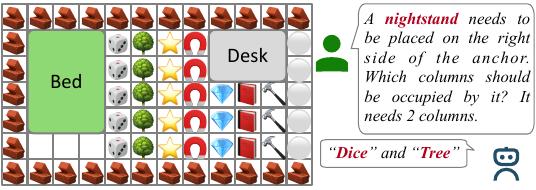}
    \caption{We discretize the top-down view into a grid and fill the grid cells with emojis. \textit{Bricks} and \textit{white go} emojis represent walls and region boundaries, respectively.}
    \label{fig:emoji-grid}
\end{figure}

To generate the position $p_l$ for the object, we use the LVLM to perceive the existing layout $C_l$ and understand the constraints defined in the proxy $P$.
In order to make the LVLM reason spatially, we propose to render the existing layout as an emoji grid.
As shown in Figure~\ref{fig:emoji-grid}, the emoji grid is a top-down projection for the layout.
Each object is denoted as a labeled rectangle.
The free spaces are filled with emojis to help the LVLM distinguish the cells.
The generation process of $p_l$ is conducted by searching for a solution in a local tree.

\noindent\textbf{Layout Node Definition.}
To make the LVLM reason $p_l$ easier, we decompose the task into three sub-tasks and enforce the LVLM to reason step by step.
Similar to the global tree, the local tree starts from an empty root node, indicating that the position is not determined yet. The first layer determines the side $\sigma_l\in\{\textrm{left},\textrm{right},\textrm{top},\textrm{bottom}\}$ relative to the anchor in top-down view.
The second layer determines the row index $u_l$.
The third layer determines the column index $v_l$.
The position is obtained from all three parameters:
\begin{equation}
    p_l=(\sigma_l, u_l, v_l).
\end{equation}

Similar to the global tree, each node in the local tree has multiple children except terminal nodes.
The terminal nodes are the ones in the last layer and those for which no valid $\sigma_l$, $u_l$, or $v_l$ exists.
If we fail to build a local tree from the root node to the last layer, the local tree solver $\pi_{\textrm{local}}$ produces $\emptyset$.

\noindent\textbf{Layout Edge Definition.}
The edge in the local tree from one step to the next is generated by an LVLM:
\begin{equation}
\sigma_l, u_l, v_l \sim \pi_{\textrm{LVLM}}(\mathcal{R}(C_l), P),
\end{equation}
where $\mathcal{R}$ renders the current layout as the emoji-grid renderer.
Here, the LVLM receives $\mathcal{R}(C_l)$ to reason visually.

\section{PRM-guided MCTS for Layout Generation}\label{sec:MCTS}
\begin{figure}
    \centering
    \includegraphics[width=.8\linewidth]{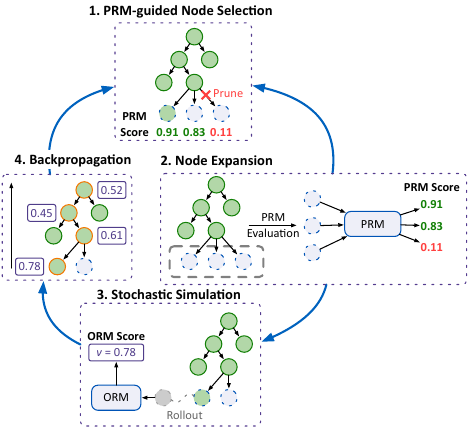}
    \vspace{-3mm}
    \caption{
    Illustration of PRM-guided MCTS for layout generation. The algorithm iteratively explores the tree-structured searching space, including node selection, node expansion, stochastic simulation, and value backpropagation.
    }
    \label{fig:prm-guided-mcts}
    \vspace{-2mm}
\end{figure}

Standard MCTS (Section~\ref{sec:formulation}) evaluates nodes solely through terminal-state scores via UCT, which prevents it from pruning unpromising branches at intermediate stages.
To address this, we propose PRM-guided MCTS, which introduces a Progress Reward Model (PRM) to score intermediate states and enable early pruning.
As shown in Figure~\ref{fig:prm-guided-mcts}, our method retains the four-step framework of standard MCTS, but we improve two steps: (1)~the \textbf{selection} step is augmented with PRM-based pruning, and (2)~the \textbf{simulation} step incorporates intermediate feedback from the PRM. The expansion and backpropagation steps remain unchanged from standard MCTS.
We describe each step below.

\noindent\textbf{(1) PRM-guided node selection.}
In standard MCTS, selection uses UCT (Eq.~\ref{eq:standard-uct}) to traverse from the root to a leaf.
We replace UCT with a PRM-augmented strategy.
Starting from the root, we iteratively select a child using the PRM-UCB criterion:
\begin{equation}
C_\textrm{select}=\mathop{\arg\max}\limits_{C'\in \textrm{children}(C)}\textrm{PRM-UCB}(C', C;\delta),
\end{equation}
where $\textrm{PRM-UCB}$ modifies the standard UCB formula by introducing a PRM-based pruning mechanism:
\begin{equation}
    \textrm{PRM-UCB}(C',C;\delta)=
\begin{cases}
\overline{v_{C'}}+\epsilon\sqrt{\frac{\ln n_C}{n_{C'}}}, & r_{C'}^{\textrm{prm}}\ge\delta\\[4pt]
-\infty, & r_{C'}^{\textrm{prm}} < \delta
\end{cases}.
\label{eq:UCB}
\end{equation}
Here, $r_{C'}^{\textrm{prm}}$ is the progress reward assigned to node $C'$ by a PRM $\mathcal{E}_{\textrm{prm}}$ (defined below), and $\delta$ is a pruning threshold.
Comparing Eq.~\ref{eq:UCB} with standard UCT (Eq.~\ref{eq:standard-uct}), the critical difference is the piecewise condition:
when $r_{C'}^{\textrm{prm}}\ge\delta$, the node is evaluated by the standard UCB formula, balancing exploration and exploitation as usual;
when $r_{C'}^{\textrm{prm}}<\delta$, the node is assigned $-\infty$, effectively pruning it from further consideration.
This mechanism prevents the search from wasting computation on branches that are already identifiable as low-quality at intermediate steps.

\noindent\textbf{(2) Node expansion.}
As in standard MCTS, the selected leaf node is expanded by generating $k$ candidate child nodes via a policy model $\pi$:
\begin{equation}
    \textrm{children}(C_{\textrm{select}}) = \{C'_i \mid C'_i\sim\pi(C_{\textrm{select}}),\; i=1,\dots,k\}.
\label{eq:children}
\end{equation}
Unlike standard MCTS, we simultaneously compute a progress reward for each newly expanded child:
\begin{equation}
    r_{C'}^{\textrm{prm}}=\mathcal{E}_{\textrm{prm}}(C', x),
\end{equation}
where $\mathcal{E}_{\textrm{prm}}$ evaluates how well the intermediate layout $C'$ satisfies the instruction $x$.
This score is used by the PRM-UCB selection criterion (Eq.~\ref{eq:UCB}) in subsequent iterations.

\noindent\textbf{(3) Stochastic simulation and (4) backpropagation.}
Steps (3) and (4) follow standard MCTS (Section~\ref{sec:formulation}): from each expanded child $C'$, we perform $n$ stochastic rollouts to terminal states and average their scores via an outcome evaluator $\mathcal{E}_{\textrm{orm}}$,
\begin{equation}
    v_{C'} = \frac{1}{n}\sum_{i=1}^{n}\mathcal{E}_{\textrm{orm}}(T^{C'}_{i}, x),
\label{eq:rollout}
\end{equation}
where $T_i^{C'}$ is a terminal state reached from $C'$ in the $i$-th rollout.
The resulting value $v_{C'}$ is backpropagated to update the ancestors of $C'$ using the standard update rule (Section~\ref{sec:formulation} Step~4).

In summary, PRM-guided MCTS retains the standard four-step MCTS framework but introduces two key improvements: (i) PRM-based pruning in the selection step, and (ii) dense intermediate feedback via progress rewards.

\section{Object Retrieval and Re-texturing}\label{sec:texture}
\begin{figure}
    \centering
    \includegraphics[width=.8\linewidth]{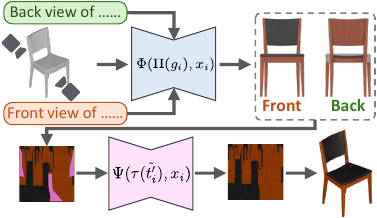}
    \vspace{-2mm}
    \caption{Illustration of our proposed texture generation approach. We use different prompts for ``front side'' and ``back side'' to alleviate the multi-face problem.}
    \vspace{-3mm}
    \label{fig:texture}
\end{figure}
In this stage, we retrieve 3D object assets from the given supporting 3D object database $\mathcal{D}$.
We retrieve the 3D asset for each object by~\cite{HoloDeck} as the following:
\begin{equation}
\begin{aligned}
\widetilde{d_i}&=\mathop{\arg\max}\limits_{d'\in\mathcal{D}} f(c_i,s_i,d'),\\
\quad f(c_i, s_i, d')&=\alpha\mathcal{C}(c_i,d')+\beta\mathcal{B}(c_i,d')+\gamma\mathcal{S}(s_i,d'),
\end{aligned}
\label{eq:retrieval}
\end{equation}

\noindent where $\mathcal{C}$ measures the CLIP similarity~\cite{CLIP} between the target category and renderings of the objects in $\mathcal{D}$,
$\mathcal{B}$ measures the SBERT similarity~\cite{SentenceBERT} between the target category and textual descriptions of the objects in $\mathcal{D}$,
$\mathcal{S}$ measures the similarities between the target size and the candidate object size.
$\alpha$, $\beta$, and $\gamma$ are trade-off hyper-parameters.

Afterwards, we generate textures for the retrieved objects. Inspired by Zeng~\etal~\cite{Paint3D}, we utilize pre-trained diffusion image generation and UV map generation models to re-texture the objects in the generated 3D scene.
For each 3D object mesh $\widetilde{d_i}$ retrieved from $\mathcal{D}$, we first generate a coarse texture map and then improve it in a refinement stage.
The 3D mesh $\widetilde{d_i}=(g_i,\widetilde{t_i})$ is decomposed into geometry $g_i$ and original texture $\widetilde{t_i}$.
We retain the geometry $g_i$ of the object and replace $\widetilde{t_i}$ with a new texture $t_i$.
Hence, the re-textured mesh is represented as $d_i=(g_i, t_i)$.

Specifically, the coarse map is first created using a depth-conditioned image diffusion model, formulated as follows:
\begin{equation}
t_i'=\Phi(\Pi(g_i), (x_i^f, x_i, x_i^b)),
\label{eq:coarse}
\end{equation}
where $\Pi$ is a camera projection function that maps the geometry $g_i$ into a depth map, $\Phi$ is a pre-trained image diffusion model that generates an image that simultaneously satisfies the depth condition $\Pi(g_i)$.
$x_i$ is the base texture description that does not contain any view constraints.
$x_i^f$ and $x_i^b$ are descriptions adding prefixes \texttt{front view of} and \texttt{back view of} to $x_i$.
To ensure view consistency and alleviate the multi-face problem, the diffusion process $\Phi$ simultaneously generates front and back views.
We use the following formulation at each sampling step~\cite{CompoDiff}:
\begin{equation}
z_j=\lambda z_j^{x_i} + (1-\lambda) \textrm{concat}(z_j^{x_i^f}, z_j^{x_i^b}),
\label{eq:texture}
\end{equation}
where $z_j^{x_i}$ is generated from the base prompt $x_i$.
It contains a consistent texture of ``front side'' and ``back side'' of the input object.
But they share the same prompt.
Thus, the pre-trained model cannot distinguish between different sides.
$z_j^{x_i^f}$ and $z_j^{x_i^b}$ are generated from $x_i^f$ and $x_i^b$ respectively. As they are prompted differently, the pre-trained model can distinguish the views.
However, the generation of $x_i^f$ and $x_i^b$ does not share context with each other.
Hence, $x_i^f$ and $x_i^b$ are not consistent.
We introduce a hyper-parameter $\lambda$ to mix $z_j^{x_i}$ and $\textrm{concat}(z_j^{x_i^f}, z_j^{x_i^b})$.
It trades off view consistency and view controllability.

During the process in Equation~\ref{eq:coarse}, where $t_i'$ is an image in the RGB space. Then, we refine $t_i'$ in the UV space with another pre-trained diffusion model:
\begin{equation}
t_i=\Psi(\tau(t_i'), x_i),
\end{equation}
where $\tau$ is a function that first warps the RGB image $t_i'$ to the mesh's surface, then unwraps to the UV space.
$\Psi$ is an inpainting model that fills the hole areas that were not generated in the first stage due to self-occlusion. 

\begin{figure*}
\centering
\subfloat[Object frequency.]{
\includegraphics[width=.37\textwidth]{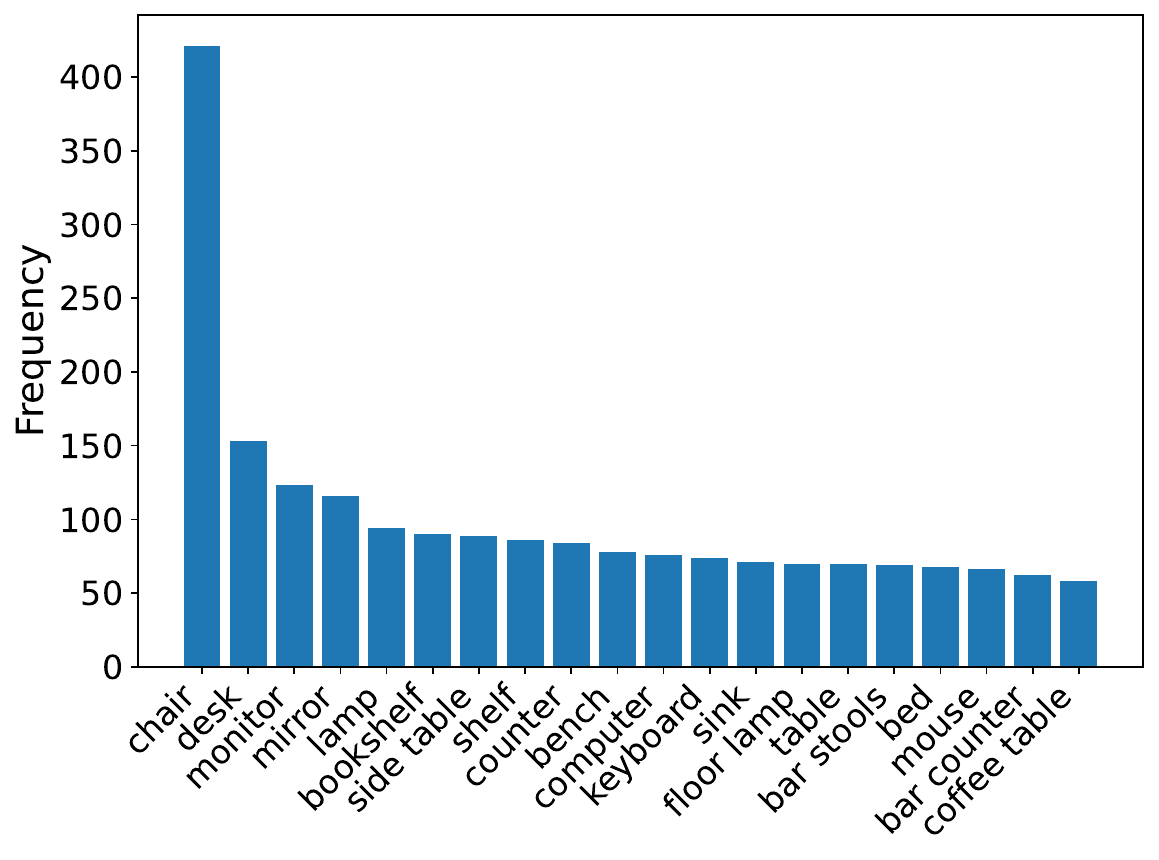}%
\label{fig:statistics-a}}
\subfloat[Word cloud for texture/style.]{
\includegraphics[width=.27\textwidth]{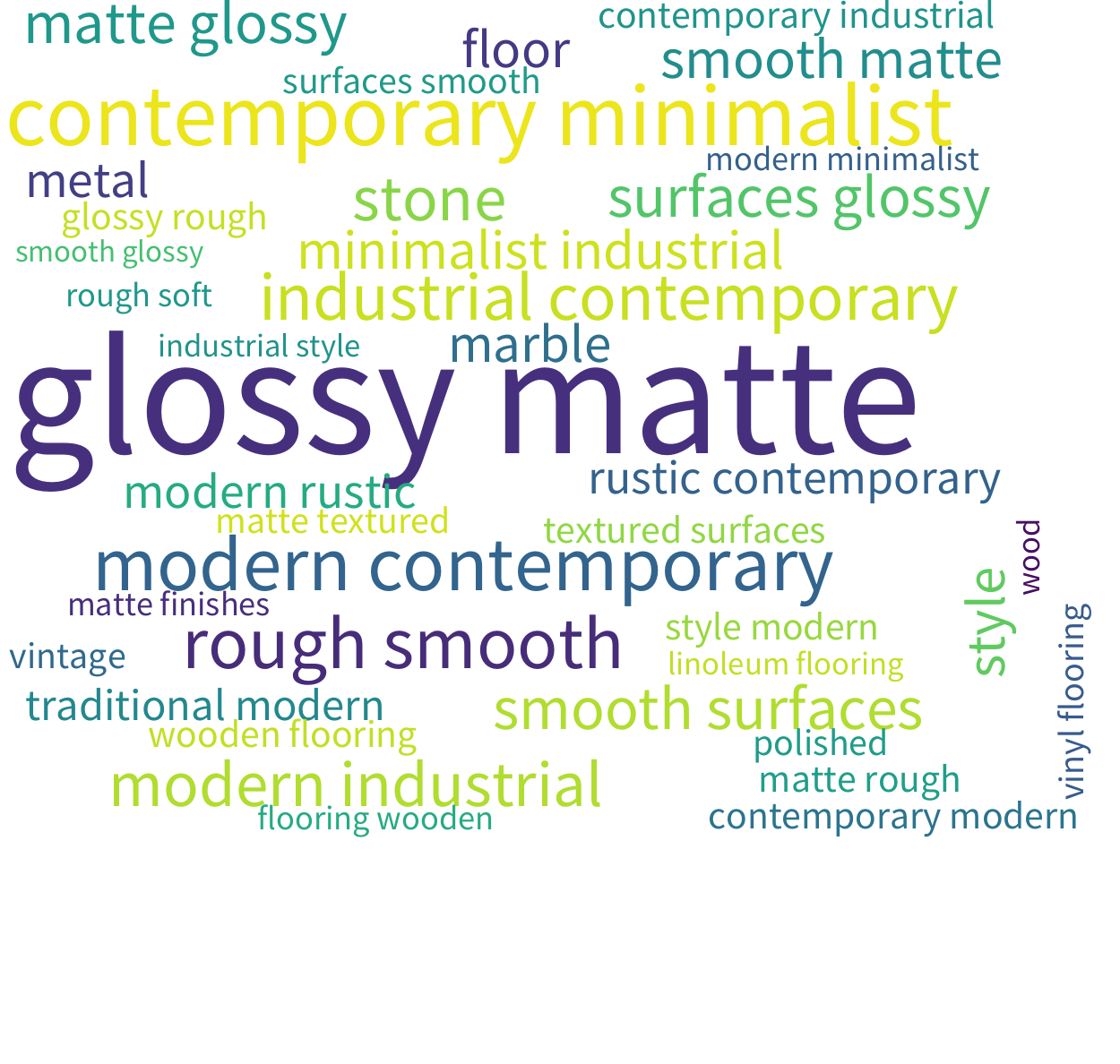}%
\label{fig:statistics-b}}
\subfloat[Inter-object relation frequency.]{
\includegraphics[width=.34\textwidth]{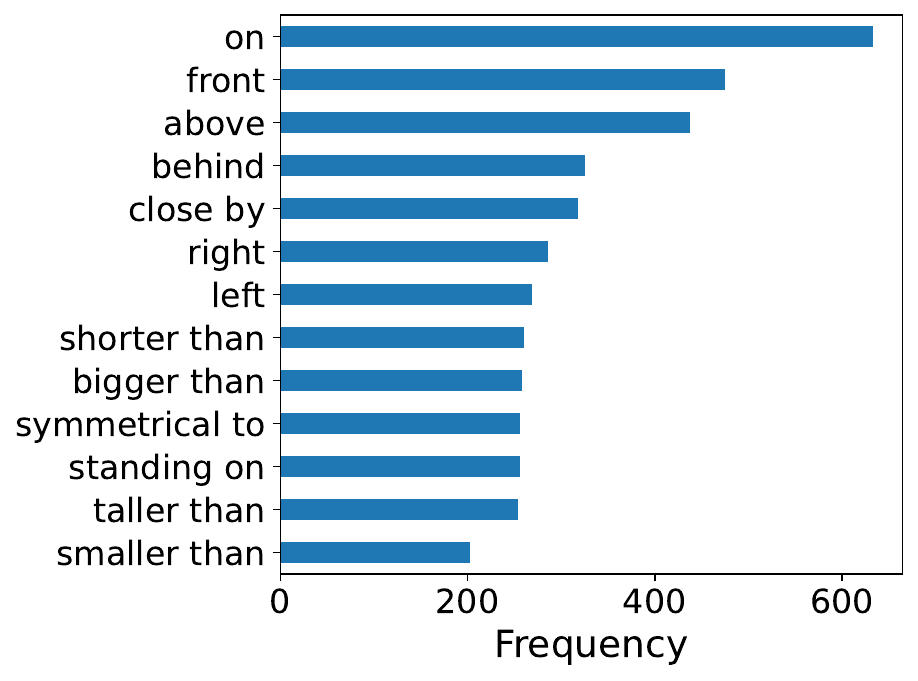}%
\label{fig:statistics-c}}
\vspace{-2mm}
\caption{Statistics of our proposed benchmark. (a) Top-20 object frequency in the textual instructions. (b) Word cloud of the texture/style requirements in the instructions. (c) Inter-object spatial relationships in the instructions.}
\vspace{-3mm}
\label{fig:statistics}
\end{figure*}

\section{Text-to-3D Indoor Scene Benchmark}

In order to fill the data gap in the task, we collect a new text-to-3D indoor scene dataset.
We aim to cover a variety of scene types, sizes, and object relationships, and provide an automatic method to assess the output scenes.

\subsection{Dataset Construction}

Our benchmark is constructed based on the common indoor scene type categorization established by the architect Neufert \etal~\cite{neufert2012architects} and MIT researchers Quattoni~\etal~\cite{MIT_Indoor_Scene}, ensuring broad applicability of our benchmark across diverse indoor environments. Specifically, we construct a text-to-3D indoor scene benchmark by prompting DeepSeek-V3 to generate textual instructions. The input conditions for each instruction are drawn from three predefined lists: \textbf{1) Scene types} --- 65 indoor scene categories (\eg, living room, kitchen, bathroom, office); \textbf{2) Room sizes} --- a set of room size descriptors (\ie, small, medium, large); \textbf{3) Inter-object relation constraints} --- 13 spatial relationships between objects. For each scene type, we sample one relation constraint and enumerate all room sizes, generating 50 instructions per scene type (3,250 in total).

Furthermore, the LLM is instructed via a system prompt to: (i) produce only factual, objective descriptions, explicitly prohibiting subjective or evaluative language; (ii) ensure logical consistency so that objects are appropriate to the scene type and arranged in realistic configurations; (iii) include hierarchical object relationships where smaller items sit on larger surfaces; and (iv) vary sentence structures to avoid repetitive patterns.
The user prompt specifies the scene type, room size, relation constraint, and target count, together with formatting requirements and diversity guidelines.
Instructions are generated in batches; when fewer than 50 valid outputs are produced for a scene type, additional batches are sampled until the target is met.

In addition, for the 3D object database $\mathcal{D}$, we use the labeled Objaverse 1.0~\cite{objaverse} processed by Yang~\etal~\cite{HoloDeck}.
They annotate the assets in Objaverse 1.0 with GPT-4V. Specifically, they input 4 orthogonal rendering images ($0^\circ$, $90^\circ$, $180^\circ$, and $270^\circ$) of an object to GPT-4V and request it to reason about the category, synset in WordNet~\cite{WordNet}, dimensions (width, length, and height in cm), volume in cm$^3$, mass in kg, which viewpoint is the frontal view, textual description, materials, placement attributes (boolean values about on\_ceiling, on\_wall, on\_floor, and on\_object).

\subsection{Dataset Statistics}

The basic statistics of our proposed benchmark are shown in Table~\ref{tab:bench-basic}.
In our benchmark, we collect 3,250 instructions for 65 scene types (50 instructions per scene type).
Compared with other datasets, ours contains more scene types, quantities, and aspects. We report detailed statistics in Figure~\ref{fig:statistics}.
As illustrated in Figure~\ref{fig:statistics-a}, most objects are \texttt{chair}, \texttt{desk}, \texttt{monitor}, and \texttt{mirror}.
They are common objects in indoor scenes.
In Figure~\ref{fig:statistics-b}, we show a word cloud for the description of texture or style for a scene.
The most frequent phrases are \texttt{glossy matte}, \texttt{modern}, \texttt{contemporary}, \texttt{contemporary minimalist}.
In Figure~\ref{fig:statistics-c}, we illustrate the frequency of spatial relationships.
The majority of relations are \texttt{on}, \texttt{front}, and \texttt{above}.
Our dataset includes diverse objects, textures/styles, and inter-object spatial relations.
It can be used for evaluating text-to-3D scene approaches comprehensively.

\begin{table}[t]
\centering
\SetTblrInner{rowsep=1pt}
\caption{
Statistics comparison of our dataset and others.
}
\vspace{-2mm}
\begin{tblr}{
colspec={cccc},
colsep={3pt},
row{1} = {font={\bfseries}},
cells={halign=c,valign=m},
hline{1,6}={1-4}{1pt, solid},
hline{2,5}={1-4}{},
}
Dataset   & \#Scene types & \#Instructions & Aspects            \\
HoloDeck~\cite{HoloDeck}  & 4             & 120            & scene type                      \\
LayoutVLM~\cite{LayoutVLM} & 11            & 33             & {scene type, layout}              \\
GALA3D~\cite{GALA3D}   & -             & 22             & layout                          \\
3DTindo-Bench      & 65            & 3,250          & {scene type, size \\layout, style}
\end{tblr}
\label{tab:bench-basic}
\vspace{-3mm}
\end{table}
\begin{center}
\footnotesize
\SetTblrInner{rowsep=1pt}
\begin{tblr}{
width=\columnwidth,
colspec={l X[1,l,m]},
colsep={3pt},
row{1} = {font={\bfseries}},
hline{1,7}={1pt, solid},
hline{2}={},
}
Metric & Description \\
Physical-Reasonability (PR) & Whether objects physically intersect or are placed outside the room boundary \\
Semantic-Reasonability (SR) & Whether objects are functionally grouped and semantically consistent with the room type \\
Layout-Reasonability (LR) & Whether objects maintain reasonable spacing and orientation \\
Instruction-Alignment (IA) & Whether the scene is consistent with the input instruction, covering object, spatial relation, and texture requirements \\
Appearance-Consistency (AC) & Whether the texture of the whole scene is visually consistent \\
\end{tblr}
\captionof{table}{Evaluation metrics for the 3DTindo-Bench, including PR, SR, LR, IA, and AC, assessed by an LVLM-based evaluator.}
\label{tab:eval-metrics}
\vspace{-3mm}
\end{center}

\subsection{3DTindo-Bench Benchmark}
\begin{figure}
    \centering
    \includegraphics[width=1.\linewidth]{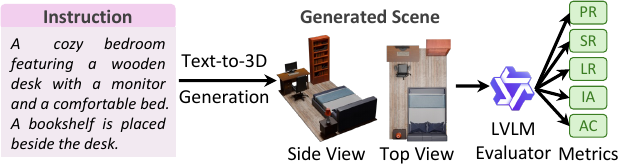}
    \vspace{-5mm}
    \caption{The pipeline for evaluation. We firstly generate 3D indoor scenes from textual instructions with different methods. Then, we render the generated 3D scenes and evaluate the renderings with a LVLM-based evaluator.}
    \label{fig:evaluation_pipeline}
\end{figure}

\begin{table*}[htbp]
\centering
\SetTblrInner{rowsep=3pt}
\caption{
Benchmark comparison with state-of-the-art text-to-3D scene generation methods.
The 3DTindo-Bench evaluates on five metrics: \textbf{PR}, \textbf{SR}, \textbf{LR}, \textbf{IA}, and \textbf{AC}.
The LayoutVLM-Bench~\cite{LayoutVLM} evaluates on: \textbf{CF}, \textbf{Pos.}, \textbf{Rot.}, and \textbf{PSA}.
We highlight the best results with \textbf{bold} and second-best with \underline{underline}.
}
\vspace{-2mm}
\begin{tblr}{
colspec={cccccccc|cccc},
colsep={3pt},
row{1-2} = {font={\bfseries}},
cells={halign=c,valign=m},
cell{1}{1,2} = {r=2}{c},
cell{1}{3} = {c=6}{c},
cell{1}{9} = {c=4}{c},
hline{1,8}={1-12}{1pt, solid},
hline{2}={3-12}{.5pt, solid},
hline{3,7}={1-12}{.5pt, solid},
}
Method & Source & 3DTindo-Bench & & & & & & LayoutVLM Benchmark~\cite{LayoutVLM} & & & \\
& & PR$\uparrow$ & SR$\uparrow$ & LR$\uparrow$ & IA$\uparrow$ & AC$\uparrow$ & Avg.$\uparrow$ & CF$\uparrow$ & Pos.$\uparrow$ & Rot.$\uparrow$ & PSA$\uparrow$ \\
HoloDeck~\cite{HoloDeck} & CVPR'24 & 97.3 & 56.7 & 52.5 & 45.1 & 60.0 & 62.3 & \underline{90.1} & \textbf{59.0} & \textbf{60.6} & \textbf{55.6} \\
LayoutVLM~\cite{LayoutVLM} & CVPR'25 & 86.0 & 54.7 & 60.5 & 47.3 & 57.6 & 61.2 & 86.7 & 34.4 & 32.1 & 25.5 \\
Deng~\etal~\cite{Global-local} & CVPR'25 & \textbf{99.9} & \underline{60.7} & \underline{70.3} & \underline{48.5} & \underline{73.4} & \underline{70.5} & \textbf{100.0} & 45.9 & \underline{52.0} & 42.4 \\
DirectLayout~\cite{DirectLayout} & NeurIPS'25 & 92.2 & 53.2 & 62.1 & 48.2 & 67.7 & 64.7 & 64.7 & 31.0 & 44.3 & 35.8 \\
Ours & -- & \underline{99.7} & \textbf{73.9} & \textbf{82.7} &\textbf{60.3} & \textbf{85.9} & \textbf{80.5} & \textbf{100.0} & \underline{53.8} & \textbf{60.6} & \underline{50.4}
\end{tblr}
\label{tab:benchmark_cmp}
\vspace{-2mm}
\end{table*}

We benchmark state-of-the-art text-to-3D indoor scene methods~\cite{HoloDeck, LayoutVLM, DirectLayout,Global-local} and our approach on our proposed dataset.
The evaluation pipeline is shown in Figure~\ref{fig:evaluation_pipeline}.
We render the 3D meshes generated by the approaches with Blender's Cycles engine in top view and side view.
The renderings and instructions generated from different methods are fairly evaluated by the same LVLM-based evaluator $\mathcal{E}$. Based on the interior design principles (Clearance, Circulation, Pairwise Relationships, Conversation, Balance, Alignment, and Emphasis) established by Merrell \etal~\cite{interior_design_guidelines}, we adopt and extend the evaluation protocol of LayoutVLM~\cite{LayoutVLM} and propose five evaluation metrics, as detailed in Table~\ref{tab:eval-metrics}.
The evaluator $\mathcal{E}$ takes the top-view and side-view renderings (concatenated as a $1024\times512$ image) and the textual instruction as input, and is instructed to provide reasons before scoring each metric from 0 to 4. All scores are normalized to $[0,100]$.

\section{Experiments}
In this section, we first introduce the experimental setup, including evaluation metrics corresponding to the 3DTindo-Bench, baselines, and implementation details.
Then we report the compared performances, ablation studies, and more analyses of our method.
\begin{figure*}
    \centering
    \includegraphics[width=0.85\linewidth]{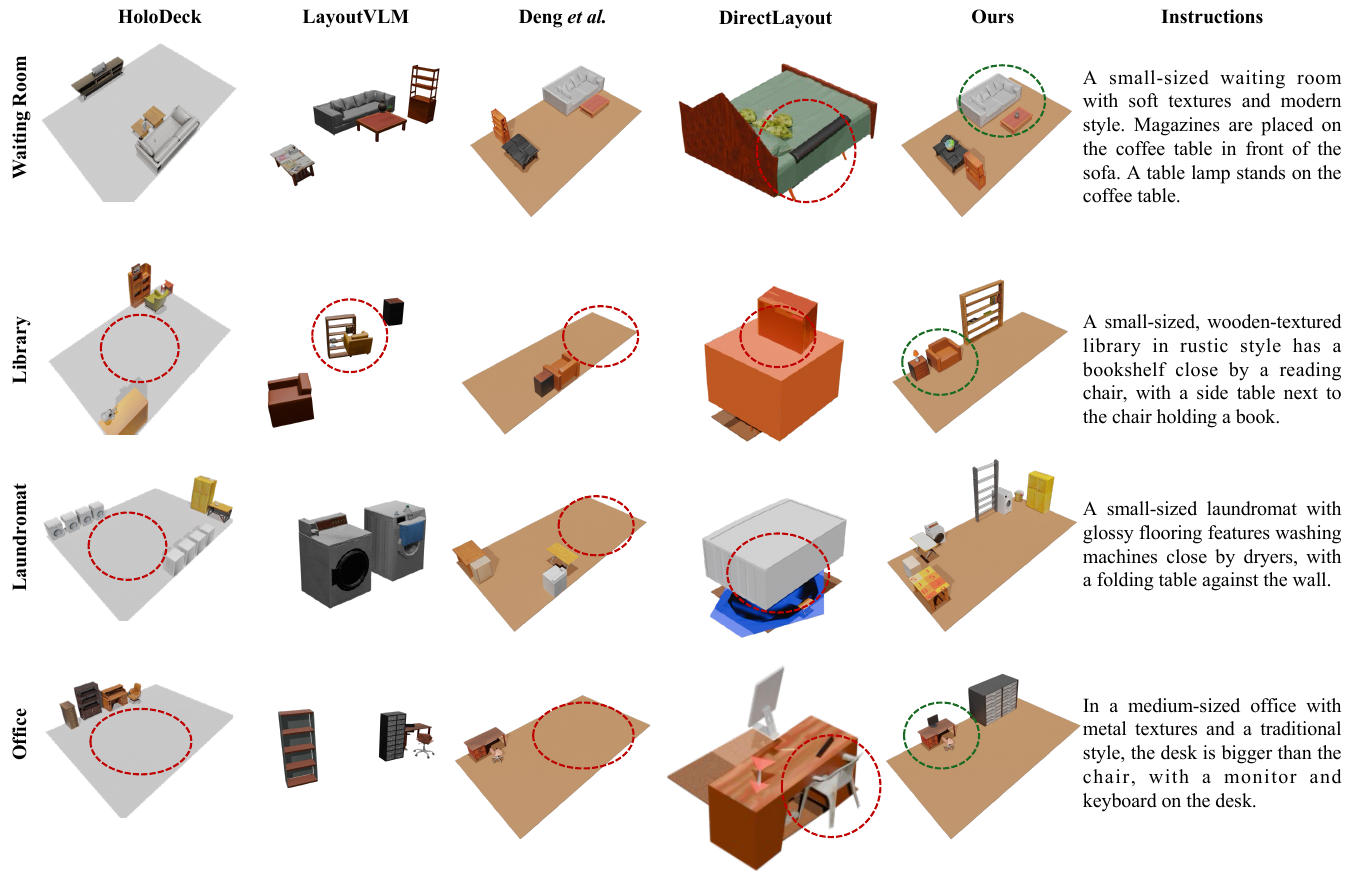}
    \vspace{-3mm}
    \caption{We visualize a number of layout generation results for \texttt{Waiting Room}, \texttt{Library}, \texttt{Laundromat}, and \texttt{Office} with HoloDeck~\cite{HoloDeck}, LayoutVLM~\cite{LayoutVLM}, Deng~\etal~\cite{Global-local}, DirectLayout~\cite{DirectLayout}, and our method.
    Non-realistic parts are marked as \textcolor{red}{red} circles and good ones are marked as \textcolor[RGB]{52,105,46}{green} circles.
    }
    \vspace{-3mm}
    \label{fig:vis_cmp}
\end{figure*}

\label{sec:experiments}
\subsection{Experimental Setup}\label{sec:exp_setup}
\noindent\textbf{Evaluation Metrics.} We evaluate our method on two benchmarks.
The first is our proposed 3DTindo-Bench, which evaluates scenes on five metrics, \ie, PR, SR, LR, IA, and AC.
The second is the LayoutVLM benchmark~\cite{LayoutVLM}, which covers 11 scene types with 3 instructions each, totaling 33 instructions, where we follow the same instruction settings as LayoutVLM and evaluate in terms of Collision-Free Score (CF), Positional Coherency (Pos.), Rotational Coherency (Rot.), and Physically-Grounded Semantic Alignment Score (PSA).
Specifically, the Collision-Free Score (CF) is adopted to evaluate the frequency of the object pairs with intersections in a scene. Positional Coherency (Pos.) and Rotational Coherency (Rot.) measure semantic alignment with the input prompt. Physically-Grounded Semantic Alignment Score (PSA) assesses both physical plausibility and semantic alignment. For Pos., Rot., and PSA, we use Qwen2.5-VL-72B as the VLM evaluator with the same instruction prompts as LayoutVLM~\cite{LayoutVLM}, scoring each layout based on its top-down and side-view renderings and the language instruction. Scores for all metrics range from $0$ to $100$, with higher scores indicating better performance.

\begin{figure*}[htbp]
\centering
\includegraphics[width=0.9\linewidth]{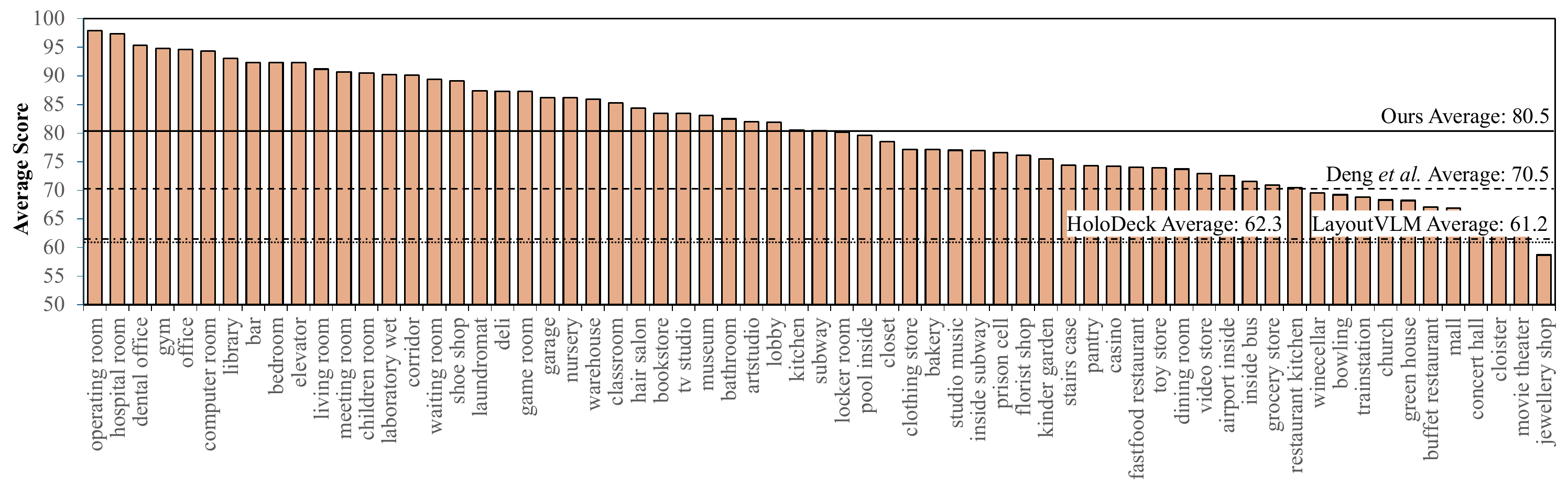}
\vspace{-2mm}
\caption{The scores of our proposed method for all 65 types in the 3DTindo-Bench from high to low.
The horizontal lines represent the performance of different methods.
}
\vspace{-3mm}
\label{fig:each_type}
\end{figure*}

\begin{figure*}[htbp]
    \centering
    \includegraphics[width=0.9\linewidth]{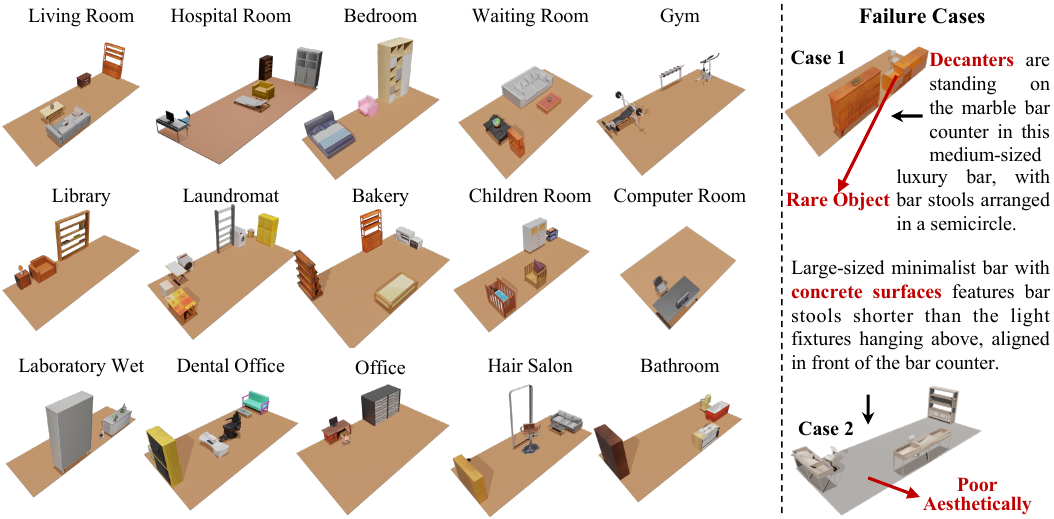}
    \vspace{-2mm}
    \caption{3D indoor layout generation results of diverse scene types from our proposed method, and our method can achieve promising performance. We also show several failure cases, including (1) rare items not in the Objaverse database cause missing item, and (2) limited aesthetics.}
    \label{fig:each_type_vis}
    \vspace{-3mm}
\end{figure*}

\noindent\textbf{Baselines.}
We compare against the following baselines: HoloDeck~\cite{HoloDeck}, LayoutVLM~\cite{LayoutVLM}, DirectLayout~\cite{DirectLayout}, and a tree search approach proposed by Deng~\etal~\cite{Global-local}\footnote{Methods published in 2026 are not included as their code is not fully available or cannot be adapted to our experimental environment.}.

\noindent\textbf{Implementation Details.} The PRM and ORM in the PRM-guided MCTS solver are implemented by using the LVLM to perceive the visually rendered results. In our experiment, we use the open-source Qwen2.5-VL-72B to generate scenes for all compared approaches and evaluate results, which is deployed via the vLLM library, and we generate scenes by calling vLLM in parallel. The exploration-exploitation trade-off hyper-parameter $\epsilon$ in Equation~\ref{eq:UCB} is set to $0.2$. We set the PRM threshold $\delta=0.3$ in Equation~\ref{eq:UCB}. The maximum number of children of a node $k$ in Equation~\ref{eq:children} is $2$. We conduct the stochastic simulation $5$ times in Equation~\ref{eq:rollout}. The maximum iteration times for the PRM-guided MCTS algorithm are 10. The retrieval function $f$ follows the design in HoloDeck~\cite{HoloDeck} and is consistent among the compared methods.
We set $\alpha=100$, $\beta=1$, and $\gamma=1$ in Equation~\ref{eq:retrieval}. For re-texturing, we set the view consistency-controllability trade-off hyper-parameter $\lambda$ in Equation~\ref{eq:texture} as $0.3$.
For the re-texturing models $\Phi$ and $\Psi$, we follow the same setting in Paint3D~\cite{Paint3D}. We use Stable Diffusion 1.5~\cite{SD} as the base model and use the ControlNet~\cite{ControlNet} to apply depth condition.

\subsection{Benchmark Performance}
\begin{table*}[htbp]
\centering
\SetTblrInner{rowsep=3pt}
\caption{
Ablation studies of our method on the 3DTindo-Bench.
\textbf{PR}=Physical-Reasonability, \textbf{SR}=Semantic-Reasonability, \textbf{LR}=Layout-Reasonability, \textbf{IA}=Instruction-Alignment, and \textbf{AC}=Appearance-Consistency.
The best results are highlighted in \textbf{bold}, and the second-best results are \underline{underlined}.
We adopt our full method and No Re-Texture as baselines, and gradually remove our innovations to validate their contributions.
The positive gains are marked as \textcolor{red}{red} and the negative drops are marked as \textcolor{blue}{blue}.
}
\vspace{-2mm}
\begin{tblr}{
colsep={4pt},
row{1} = {font={\bfseries}},
hline{2,4}={1-8}{.5pt, solid},
hline{1,9}={1-8}{1pt, solid},
cell{1,2,3,4,5,6,7,8}{2,3,4,5,6,7,8}={c},
cell{1}{1}={c},
}
Ours & PR$\uparrow$ & SR$\uparrow$ & LR$\uparrow$ & IA$\uparrow$ & AC$\uparrow$ & Avg.$\uparrow$ & Avg. Time Consumption$\downarrow$ \\
Ours & \textbf{99.7} & \textbf{73.9} & \textbf{82.7} & \textbf{60.3} & \textbf{85.9}& \textbf{80.5} & -- \\
$\quad$w/o   Re-Texture & \underline{99.6}(\textcolor{blue}{-0.1}) & \underline{73.4}(\textcolor{blue}{-0.5}) & \underline{81.8}(\textcolor{blue}{-0.9}) & \underline{60.1}(\textcolor{blue}{-0.2}) & \underline{84.0}(\textcolor{blue}{-1.9}) & \underline{79.7}(\textcolor{blue}{-0.8})  & -- \\
No Re-Texture Baseline & 99.6 & \textbf{73.4} & \textbf{81.8} & \textbf{60.1} & -- & \textbf{78.7}  & \textbf{$\sim$ 1.87 Min.} \\
$\quad$w/o Tree Search & 99.5(\textcolor{blue}{-0.1}) & 71.0(\textcolor{blue}{-2.4}) & 80.4(\textcolor{blue}{-1.4}) & 57.9(\textcolor{blue}{-2.2}) & -- & 77.2(\textcolor{blue}{-1.5})  & -- \\
$\quad$w/o Visual Rendering & \underline{99.8}(\textcolor{red}{+0.2}) & 69.6(\textcolor{blue}{-3.8}) & 79.4(\textcolor{blue}{-2.4}) & 56.6(\textcolor{blue}{-3.5}) & -- & 76.3(\textcolor{blue}{-2.4})  & -- \\
$\quad$w/o PRM & 99.5(\textcolor{blue}{-0.1}) & \underline{72.8}(\textcolor{blue}{-0.6}) & \underline{81.4}(\textcolor{blue}{-0.4})  & \underline{59.6}(\textcolor{blue}{-0.5}) & -- & \underline{78.3}(\textcolor{blue}{-0.4})  &$\sim$ 2 Min.(\textcolor{blue}{+0.13}) \\
$\quad$w/o MCTS & \textbf{99.9}(\textcolor{red}{+0.3}) & 60.7(\textcolor{blue}{-12.7}) & 70.3(\textcolor{blue}{-11.5}) & 48.5(\textcolor{blue}{-11.6}) & -- & 69.8(\textcolor{blue}{-8.9}) & -- 
\end{tblr}
\label{tab:ablation_cmp}
\end{table*}

Our method outperforms state-of-the-art methods as reported in Table~\ref{tab:benchmark_cmp} (left) on the 3DTindo benchmark and we show several samples generated by different methods in Figure~\ref{fig:vis_cmp}.
Averaging over 3,250 scenes and 5 metrics, our method improves by 10.0 scores compared to the best-performing existing baseline~\cite{Global-local}.
We also achieve the best among almost all metrics.
HoloDeck performs well in PR due to its hard constraint in physics.
However, it struggles to generate realistic scenes due to its human-defined constraints, which are not perfect for finding a global optimal layout, resulting in large space gaps in scenes as shown in Figure~\ref{fig:vis_cmp}.
LayoutVLM highly relies on reducing human-defined losses and on the initial layout yielded by pre-trained LVLMs.
As illustrated in Figure~\ref{fig:vis_cmp}, LayoutVLM generates a \texttt{Library} with object intersections.
LayoutVLM performs relatively poorly because it highly relies on the spatial knowledge of pre-trained LVLMs and human-defined placement rules.
The method proposed by Deng~\etal achieves good results in the metrics due to its tree-search-based modeling, but it fails to generate the most realistic scenes because it uses a greedy DFS algorithm.
However, the DFS algorithm does not score intermediate states, which makes the algorithm skip a local optimum and miss some assets.
Notably, our approach utilizes an MCTS-based algorithm for layout generation and re-textures the objects, thereby maintaining high realism in both layout and appearance.

Besides comparing on our 3DTindo-Bench, we also evaluate on the LayoutVLM benchmark~\cite{LayoutVLM}.
The results are reported alongside the 3DTindo-Bench results in Table~\ref{tab:benchmark_cmp} (right).
On the CF score, our method achieves a perfect 100.0, tying with Deng~\etal~\cite{Global-local}.
Both methods formulate layout generation as a tree-structured search.
When an object placement causes intersections, the algorithm can backtrack to earlier nodes and explore different spatial configurations, effectively correcting placement errors.
In contrast, LayoutVLM (86.7) and DirectLayout (64.7) place objects sequentially without backtracking, so once a collision occurs it propagates and cannot be corrected.
HoloDeck (90.1) mitigates collisions via hard constraints but still falls short of tree-search-based methods in crowded scenes.
For Pos., our method ranks second (53.8) behind HoloDeck (59.0), followed by Deng~\etal (45.9), LayoutVLM (34.4), and DirectLayout (31.0).
Both tree-search-based methods (ours and Deng~\etal) leverage the LVLM's spatial prior to select object positions, achieving competitive results without hand-crafted rules.
Our MCTS-based search further outperforms Deng~\etal's DFS by better balancing exploration and exploitation, yielding higher positional coherence.
While HoloDeck's predefined rules achieve the highest Pos., they are prone to overfitting the LayoutVLM benchmark, which evaluates only 11 scene types, allowing rule-based methods to excel in these specific categories but limiting their generalization to diverse scenes.
For Rot., our method (60.6) ties with HoloDeck (60.6) as the best, exceeding Deng~\etal (52.0), DirectLayout (44.3), and LayoutVLM (32.1).
This demonstrates that our step-by-step MCTS placement, guided by the LVLM's semantic understanding, achieves precise orientation alignment without relying on hand-crafted rules.
For PSA, which is a holistic VLM evaluation of both physical plausibility and semantic alignment, our method ranks second (50.4) behind HoloDeck (55.6).
HoloDeck's first place relies on predefined rules that excel on the limited 11 scene types of the LayoutVLM benchmark, but such rule-based approaches lack generalization to diverse scenes.

We also report the average benchmark score of our method on each scene type in Figure~\ref{fig:each_type} and illustrate the layout generation results on diverse types of scenes in Figure~\ref{fig:each_type_vis}.
Our method can generalize to diverse scene types since we utilize a well-trained LVLM.
Moreover, our approach can generate plausible results on more scene types compared with existing solutions.
However, as shown in Figure~\ref{fig:each_type_vis} (right), our method has several limitations:
(1) when prompts specify rare or niche objects (such as \texttt{decanters}) that are not available in the Objaverse database, the corresponding items fail to appear in the generated scene, as the retrieval function cannot find matching 3D models (\textit{e.g.}, expanding the 3D database or incorporating generative 3D models could mitigate this issue);
(2) the aesthetic quality of generated textures sometimes falls short of human expectations, as the LVLM's understanding of design principles and style coherence may not fully capture nuanced aesthetic preferences (\eg, incorporating aesthetic reward models or style-consistent generative priors could improve visual quality). 
These limitations stem from the reliance on external 3D model databases and the current capabilities of large vision-language models in understanding and generating visually appealing content.

\subsection{Ablation Study}

\begin{figure*}
\centering
\includegraphics[width=0.9\linewidth]{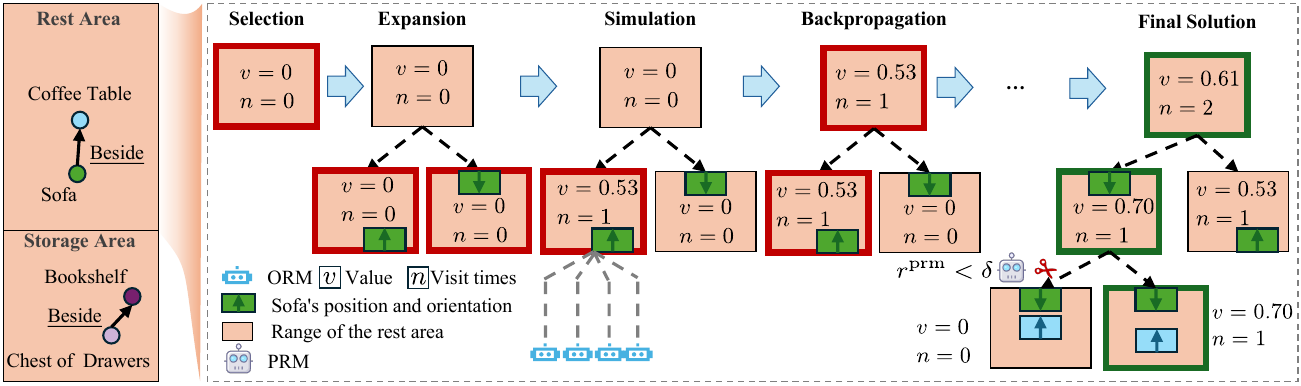}
\vspace{-2mm}
\caption{This figure demonstrates the layout generation process of our proposed method. We first generate a hierarchical scene representation (left), then we utilize the PRM-guided MCTS algorithm, including selection, expansion, simulation, and backpropagation processes, to get a final solution (right).}
\vspace{-2mm}
\label{fig:intermediate_steps}
\end{figure*}

We conduct ablation studies on the 3DTindo-Bench to assess the key components, and results are reported in Table~\ref{tab:ablation_cmp}.

\noindent\textbf{Effectiveness of texture generation.}
Our full method applies re-texturing to all objects for appearance consistency, achieving AC score of 85.9 and an average score of 80.5.
In the w/o Re-Texture variant, we use original textures of objects retrieved from Objaverse, which yields an AC score of 84.0 and an average score of 79.7, dropping by 1.9 and 0.8 points, respectively.
This indicates that re-texturing effectively improves the global appearance coherence without compromising layout quality.

\noindent\textbf{Effectiveness of tree search.}
Our full method models layout generation as a tree search, enabling backtracking when a placement error occurs.
In the w/o Tree Search variant, we replace the tree-structured modeling with a chain-structured one that places objects sequentially without branching.
The average score drops from 78.7 to 77.2 ($-1.5$), with SR decreasing by 2.4 (73.4$\to$71.0) and IA by 2.2 (60.1$\to$57.9).
This is because the chain structure cannot recover from placement errors: once an object with key semantics for the room is misplaced, subsequent objects are also affected and cannot be corrected.

\noindent\textbf{Effectiveness of visual rendering.}
Our full method renders the current layout as an emoji grid to provide spatial feedback to the LVLM.
In the w/o Visual Rendering variant, we represent the existing layout with a free-form textual description instead.
The average score drops from 78.7 to 76.3 ($-2.4$), with SR decreasing by 3.8 (73.4$\to$69.6), LR by 2.4 (81.8$\to$79.4), and IA by 3.5 (60.1$\to$56.6).
This demonstrates that visual prompts convey spatial information more effectively than text alone for layout reasoning.

\noindent\textbf{Effectiveness of the PRM.}
Our full method employs a PRM to evaluate intermediate states and prune unpromising branches during selection.
In the w/o PRM variant, we replace the PRM-guided selection with standard UCT selection that only uses terminal-state scores.
The average score changes marginally from 78.7 to 78.3 ($-0.4$), while the generation time increases from 1.87 to 2.00 minutes ($+0.13$).
This confirms that the PRM improves search efficiency without sacrificing solution quality. Theoretically, the efficiency gain comes from the PRM-guided selection step.
Let $d$ be the tree depth and $b$ the average branching factor.
Standard MCTS evaluates all $b$ children at each level via UCT, with a selection complexity of $O(d \cdot b)$.
In PRM-guided MCTS, nodes whose PRM score falls below the threshold $\delta$ are pruned (assigned $-\infty$), reducing the effective branching factor to $b' = p_{\delta} \cdot b$, where $p_{\delta} = P(r_{C'}^{\textrm{prm}} \ge \delta)$.
The selection complexity becomes $O(d \cdot p_{\delta} \cdot b)$.

\noindent\textbf{Effectiveness of MCTS.}
Our full method uses PRM-guided MCTS to search the layout tree with exploration--exploitation balancing and intermediate-state evaluation.
In the w/o MCTS variant, we replace MCTS with a greedy DFS algorithm that commits to the first viable placement without backtracking.
The average score plummets from 78.7 to 69.8 ($-8.9$), with severe drops across SR (73.4$\to$60.7, $-12.7$), LR (81.8$\to$70.3, $-11.5$), and IA (60.1$\to$48.5, $-11.6$).
This is because DFS lacks the ability to evaluate intermediate states and backtrack from poor placements, making it prone to local optima.

\subsection{More Analysis}
\begin{figure}
    \centering
    \includegraphics[width=0.9\linewidth]{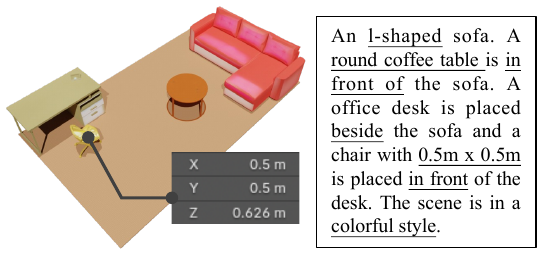}
    \caption{Visualization demonstrating that our proposed method can control attributes given in a complex instruction. The instruction constrains ``L-shaped sofa'', ``round coffee table'', ``in front of'', ``beside'', ``0.5m x 0.5m'', ``colorful style''. Our method achieves these constraints well.}
    \vspace{-3mm}
    \label{fig:attribute_control}
\end{figure}

\begin{figure*}
\centering
\includegraphics[width=0.8\linewidth]{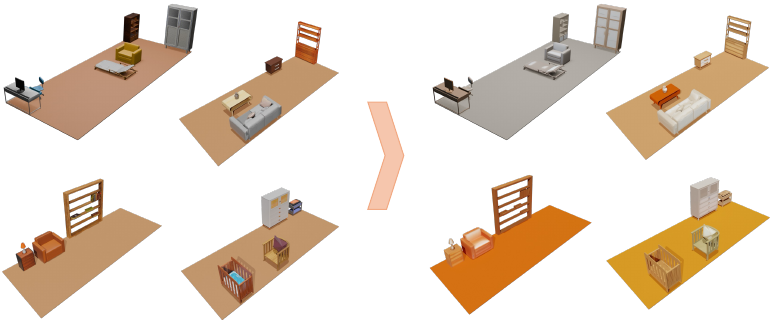}
\caption{Applying our re-texture method to a scene benefits the appearance consistency. (Left) Before re-texturing. (Right) After re-texturing.}
\label{fig:re-texture-full}
\end{figure*}

\begin{figure*}
    \centering
    \includegraphics[width=.7\linewidth]{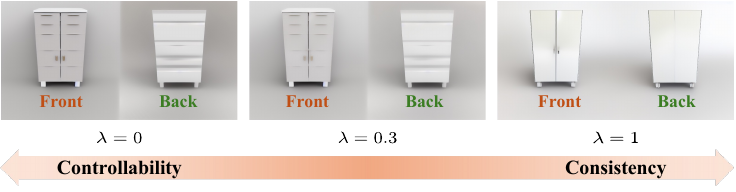}
    \vspace{-2mm}
    \caption{
    The coarse maps generated in the first stage of the re-texture module with different $\lambda$ settings.
    In each setting, our proposed method generates the \texttt{front} and \texttt{back} views for an object.
    With less $\lambda$, the controllability increases, but the appearances for the views are less consistent.
    While along with more $\lambda$, the view consistency increases, but the controllability decreases.
    Then we set $\lambda=0.3$ between $0$ and $1$ to trade off between them.
    }
    \vspace{-3mm}
    \label{fig:re-texture-trade-off}
\end{figure*}

\noindent\textbf{Intermediate Steps.}
Figure~\ref{fig:intermediate_steps} illustrates an example of how we generate a layout step-by-step.
First, we use an LVLM to generate an intermediate scene representation.
It decomposes a scene into a \texttt{rest area} and a \texttt{storage area}.
Based on the representation, we conduct the PRM-guided MCTS algorithm to generate a layout.
Specifically, the initial problem tree consists of a single root node, indicating an empty layout.
The value $v$ and visit times $n$ are $0$ currently.
The algorithm \textbf{selects} the node and \textbf{expands} two child nodes with Equation~\ref{eq:children}.
After that, we choose one child to perform a stochastic \textbf{simulation}.
The simulated results are evaluated with the ORM, yielding a value $v=0.53$ and increasing the visit times to $1$.
Following, the algorithm performs a \textbf{backpropagation} from the simulated node to the root node, updating the value and visit times of the root node.
The selection-expansion-simulation-backpropagation iteration runs multiple times.
During the process, we use a PRM to validate if a node can be pruned.
Finally, we get a route from the root to a terminal, indicating a solution.

\noindent\textbf{Scene Control.}
Figure~\ref{fig:attribute_control} shows a scene generated from a complex instruction.
We instruct our model with specific object shape, dimensions, relationships, and style constraints.
From the results, we can infer that our method can accomplish complex user requirements.
This capability is derived from the well-trained LVLM, which can understand and follow user requirements.

\noindent\textbf{Texture Generation.}
Figure~\ref{fig:re-texture-full} shows the advantage of our re-texture method.
Before re-texturing (left), the global appearance is not consistent.
For example, in the first scene, the \texttt{chair} before the desk is \texttt{light blue} while the \texttt{cabinet} at the corner is \texttt{dark brown}.
After applying our re-texture method (right), the objects in the scene share a coherent appearance.
That is because we use an LVLM to reason about the style of each object with a global context. Figure~\ref{fig:re-texture-trade-off} illustrates that our method can trade off the view controllability and consistency during texture generation.
Specifically, when we set $\lambda=0$ in Equation~\ref{eq:texture}, the textures for the frontal and back views are independently generated with prompts \texttt{a frontal view of a cabinet} and \texttt{a back view of a cabinet}, respectively.
The front texture contains more patterns which belong to the frontal side of a cabinet, \eg, \texttt{the handles}.
However, the frontal texture is darker than the back texture because they do not share context with each other.
In contrast, when we set $\lambda=1$ in Equation~\ref{eq:texture}, the textures are generated with the same prompt.
In Figure~\ref{fig:re-texture-trade-off} (right), we can observe that the two-sided textures are more consistent, but they contain fewer patterns which should appear on the frontal side.
To this end, we set $\lambda$ between $0$ and $1$, which makes a trade-off between controllability and consistency.

\section{Conclusion}
\label{sec:conclusion}
In this paper, we presented a new reasoning method to leverage a pre-trained LVLM for 3D indoor scene generation via a training-free approach.
To bridge the semantic gap between free-form instructions and scenes, we abstracted a scene with a hierarchical representation and incorporated it with a PRM-guided MCTS algorithm.
The PRM pruned the searching tree and the MCTS algorithm efficiently explored the searching space, which significantly boosts the reasoning process of LVLMs.
Furthermore, we introduced a new benchmark, named 3DTindo-Bench, to boost text-to-3D indoor scene generation.
Extensive experimental results on 3DTindo-Bench demonstrated that our method outperformed state-of-the-art methods and produced diverse and realistic results.
We will extend the method to robot navigation and embodied AI in the future.

\nocite{SGFormer,lv_neurips,qi_tip,STC-GAN,qi_CVPR19,qi_tcsvt,RDFC-GAN}
\nocite{DC-SAM,LvTPAMI,YeTIP,PITN,SafeDriveRAG,T2SG,ZhuICCV,STC-GAN,qi_CVPR19,Robo-SGG}

\bibliographystyle{IEEEtran_etal}
\bibliography{main}

\clearpage
\appendices

\appendices
\section{Implementation Details}
In this section, we introduce the details of generating the hierarchical scene representation and the thought tree definition respectively.
\subsection{Hierarchical Scene Representation}
To generate the hierarchical scene representation from textual input from users, $p^{x\to P}$, we sequentially generate each level in the representation with the VLM, \ie, room, region, floor object, and supported object levels.

\noindent\textbf{Room-level generation.}
We prompt the VLM to estimate the dimensions with the prompt template as shown in Figure~\ref{fig:get-room-node}.

\noindent\textbf{Region-level generation.}
We pre-define several functional regions in the prompts to generate the functional regions for different types of rooms with the VLM.
The prompt template is shown in Figure~\ref{fig:get-region-node}.
Then, we use the prompt in Figure~\ref{fig:get-region-layout} to estimate the length for each region and arrange them.

\noindent\textbf{Floor-object-level generation.}
For each region, we first prompt the VLM to generate the object set $S'=\{o_1',o_2',\cdots,o_N'\}$ in the floor object level with the template prompt shown in Figure~\ref{fig:get-floor-objects}.
During this process, we enforce the VLM to generate at least one primary object that represents the main function of a region in a room (\eg, a \textit{bed} is essential for the \textit{rest region} of a \textit{bedroom}).
In addition, the VLM is required to estimate the dimensions of each object and decide which side (longer or shorter) tends to be the frontal side of the object (\eg, longer for sofas, shorter for beds, \etc).
In Figure~\ref{fig:get-anchor} and Figure~\ref{fig:get-anchor-placement}, we leverage the VLM to choose an anchor object and determine the placement rule for it.
Second, we prompt the VLM with the template in Figure~\ref{fig:get-affiliated-placement} to determine the edge set $E$.
The spatial relationship contains the placement rules (\ie \textit{place\_front}, \textit{place\_beside} and \textit{place\_around}) and alignment rules (\ie \textit{side\_alignment}, \textit{center\_alignment}, and \textit{around}).
The alignment rules are used to help the row or column determination steps.
For example, an \textit{sofa} and an \textit{TV stand} are usually center aligned.
They share the same row or column centers and can be directly calculated.

\noindent\textbf{Supported-object-level generation.}
In this layer, we share similar prompts with the floor-object-level generation.

\subsection{Tree Thought Definition}
\begin{figure}
    \centering
    \includegraphics[width=.4\linewidth]{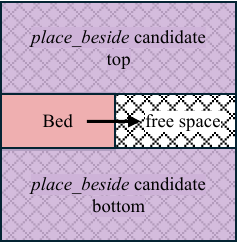}
    \caption{An example of calculating side candidates.}
    \label{fig:side-example}
\end{figure}

\noindent\textbf{Side determination.}
We first calculate the side candidates based on the anchor's location and orientation, and the placement rule of the current object.
For example, as shown in Figure~\ref{fig:side-example}, if a \textit{bed} is face to the right and an \textit{nightstand} need to be \textit{place beside} of the \textit{bed}, there are two candidates: \textit{top} and \textit{bottom}. Then, we prompt the VLM to choose a reasonable side from the candidates, as shown in Figure~\ref{fig:side-determination}.

\noindent\textbf{Row/Columns determination.}
We use the prompt template in Figure~\ref{fig:row-column-determination} to choose emojis, denoting the rows or columns.
The VLM is required to output \textit{``None''} if it finds no suitable rows or columns to put the object, which serves as the thought evaluator for the previous step.

\begin{figure*}
    \centering
    \includegraphics[width=1\linewidth]{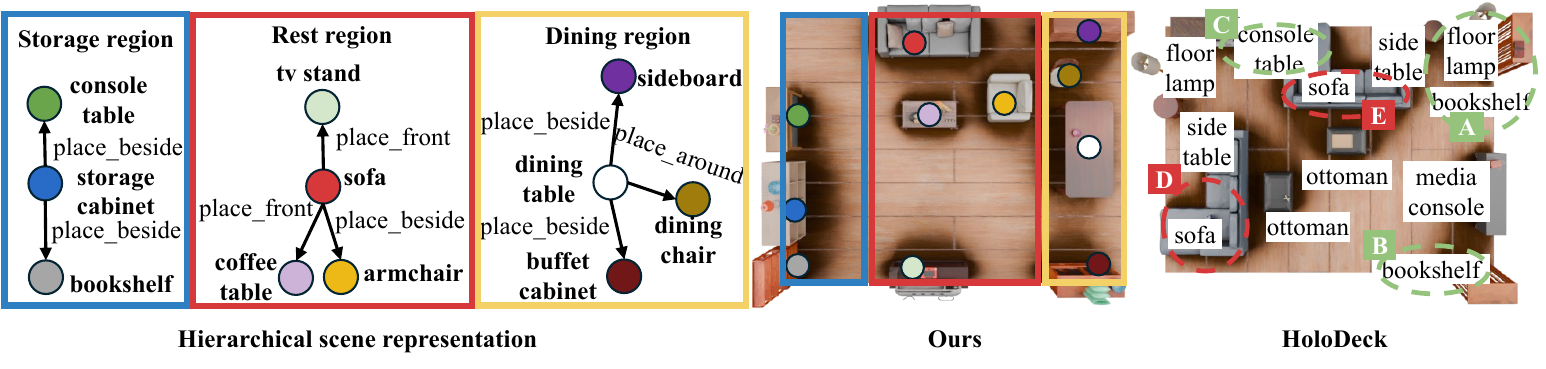}
    \caption{
    Compared with HoloDeck, our method's result exhibits a clearer division of functional regions and improved spatial harmony.
    }
    \label{fig:hierarchy-effective}
\end{figure*}

\subsection{Algorithm of DFS-based and MCTS-based Scene Generation}
\noindent\textbf{DFS}: The DFS-based tree search method is shown in Algorithm~\ref{algo:dfs}.
\begin{algorithm}[!ht]
    \renewcommand{\algorithmicrequire}{\textbf{Input:}}
    \renewcommand{\algorithmicensure}{\textbf{Output:}}
    \caption{DFS algorithm for global and local tree search}
    \label{algo:dfs}
    \begin{algorithmic}[1]
        \Require Thought generator G, current layer $i$, maximum attempt $k$, maximum layer $I$, current tree state $s_i$, edge set $E$
        \Ensure Success/Failure
        
        \If {$i>I$}
            \State \Return True
        \EndIf
        \For {$i = 1 \cdots k$}
            \If {$s_{i+1} \sim G(s_i, e_{i,a}\in E$) \textbf{and} \textsc{DFS}($i+1$,$s_{i+1}$,$e_{i+1,a}\in E$)}
                \State \Return True
            \EndIf
        \EndFor
        
        \State \Return False
    \end{algorithmic}
\end{algorithm}

\noindent\textbf{MCTS}: The PRM-guided MCTS-based tree search method is shown in Algorithm~\ref{algo:mcts}.
\begin{algorithm}[!ht]
    \renewcommand{\algorithmicrequire}{\textbf{Input:}}
    \renewcommand{\algorithmicensure}{\textbf{Output:}}
    \caption{PRM-guided MCTS algorithm for global and local tree search}
    \label{algo:mcts}
    \begin{algorithmic}[1]
        \Require Root node $C_0$, policy model $\pi$, progress reward model $\mathcal{E}_{\textrm{prm}}$, outcome reward model $\mathcal{E}_{\textrm{orm}}$, instruction $x$, threshold $\delta$, max children $k$, max simulations $n$, exploration $\epsilon$, max iterations $T$
        \Ensure Trajectory from $C_0$ to terminal node
        
        \For{$iter = 1 \cdots T$}
            \State \textbf{Step 1: Selection}
            \State $C_{\textrm{select}} \gets C_0$
            \While{$C_{\textrm{select}}$ is not a leaf node}
                \State $C_{\textrm{select}} \gets \displaystyle\mathop{\arg\max}\limits_{C'\in \textrm{children}(C)} \textrm{PRM-UCB}(C', C; \delta)$
            \EndWhile
            \State
            \State \textbf{Step 2: Expansion}
            \State $\textrm{children}(C_{\textrm{select}}) \gets \{C'_i | C'_i \sim \pi(C_{\textrm{best}}), i=1,\cdots,k\}$
            \For{each $C' \in \textrm{children}(C_{\textrm{select}})$}
                \State $r_{C'}^{\textrm{prm}} \gets \mathcal{E}_{\textrm{prm}}(C', x)$
            \EndFor
            \State
            \State \textbf{Step 3: Stochastic simulation}
            \For{each $C' \in \textrm{children}(C_{\textrm{select}})$}
                \State $v_{C'} \gets 0$
                \For{$sim = 1 \cdots n$}
                    \State $T^{C'}_{sim} \gets \textsc{Simulate}(C')$
                    \State $v_{C'} \gets v_{C'} + \mathcal{E}_{\textrm{orm}}(T^{C'}_{sim}, x)$
                \EndFor
                \State $v_{C'} \gets v_{C'} / n$
            \EndFor
            \State
            \State \textbf{Step 4: Value backpropagation}
            \For{each $C' \in \textrm{children}(C_{\textrm{select}})$}
                \For{each node $C_i$ in path from $C_0$ to $C'$}
                    \State $v_{C_i} \gets v_{C_i} + v_{C'}$
                    \State $n_{C_i} \gets n_{C_i} + 1$
                \EndFor
            \EndFor
            \State
            \If{$\exists C' \in \textrm{children}(C_{\textrm{select}})$ is terminal node}
                \State \Return path from $C_0$ to $C'$
            \EndIf
        \EndFor
        \State \Return best path from $C_0$ to terminal node based on average value
    \end{algorithmic}
\end{algorithm}

\section{More Comparisons}
In this section, we showcase more comparison results to demonstrate the effectiveness of our method.

\subsection{Parameter Study}

We conduct a parameter study to analyze the sensitivity of the PRM-guided MCTS solver to its key hyper-parameters.
Specifically, we study three parameters: 
1) the exploration-exploitation trade-off $\epsilon$ (Equation~(11)), which balances exploring less-visited nodes against exploiting high-value nodes. 
A larger $\epsilon$ encourages more exploration, helping to avoid local optima at the cost of more search steps.
2) the PRM threshold $\delta$ (Equation~(11)), which determines the pruning criterion for low-quality nodes.
A higher $\delta$ prunes more nodes, reducing the branching factor but potentially discarding promising candidates.
3) the maximum number of children $k$ (Equation~(12)), which controls how many candidate child nodes are expanded at each MCTS iteration.
A larger $k$ broadens the search space but increases computational cost.

\begin{table}[htbp]
\centering
\SetTblrInner{rowsep=3pt}
\caption{
Parameter study of hyper-parameters on the 3DTindo-Bench.
We study three hyper-parameters: the exploration-exploitation trade-off $\epsilon$ (ours: 0.2), the PRM threshold $\delta$ (ours: 0.3), and the maximum number of children of a node $k$ (ours: 2).
Each parameter is varied independently while keeping others unchanged.
}
\vspace{-2mm}
\begin{tblr}{
colspec={llcccccc},
colsep={3pt},
row{1} = {font={\bfseries \footnotesize}},
rows = {font=\footnotesize},
cells={halign=c,valign=m},
cell{1}{1,2} = {c},
hline{2,5}={1-8}{.5pt, solid},
hline{1,6}={1-8}{1pt, solid},
}
Param. & Setting & PR$\uparrow$ & SR$\uparrow$ & LR$\uparrow$ & IA$\uparrow$ & AC$\uparrow$ & Avg.$\uparrow$ \\
$\epsilon$ & 0.4 & \textbf{99.9} & 71.0 & 80.0 & 58.0 & 82.8 & 78.3 \\
$\delta$ & 0.6 & \underline{99.8} & 71.1 & 80.0 & 58.1 & 83.1 & 78.4 \\
$k$ & 3 & 99.7 & \underline{73.0} & \underline{81.2} & \underline{59.6} & \underline{84.5} & \underline{79.6} \\
Ours & $\epsilon=0.2$, $\delta=0.3$, $k=2$ & 99.7 & \textbf{73.9} & \textbf{82.7} & \textbf{60.3} & \textbf{85.9} & \textbf{80.5} \\
\end{tblr}
\label{tab:param_study}
\end{table}

From the Avg. score in Table~\ref{tab:param_study}, we observe that our default configuration achieves the best overall performance (80.5). 
Raising $\epsilon$ from 0.2 to 0.4 decreases Avg. to 78.3. From the MCTS perspective, a larger $\epsilon$ increases the exploration bonus in the UCB formula, which diverts the search toward less visited branches and away from exploiting the most promising nodes, ultimately wasting the limited iteration budget on low-reward trajectories.
Increasing $\delta$ from 0.3 to 0.6 drops Avg. to 78.4. 
Compared with Ours, a higher PRM threshold leads to more aggressive pruning of candidate nodes during tree expansion.
While this reduces the branching factor, it also risks eliminating viable layouts early, thereby narrowing the effective search space and preventing the solver from reaching high-quality solutions. 
Increasing $k$ from 2 to 3 yields an Avg. of 79.6, the closest to our default.
From the MCTS perspective, a larger $k$ expands more child nodes per iteration, which broadens the search at each step but also increases the branching factor, potentially diluting the search depth under a fixed iteration budget and leading to marginally lower overall quality.

\noindent\textbf{Effective of the hierarchical scene representation.}
We compare our method with HoloDeck in Figure~\ref{fig:hierarchy-effective}.
In this figure, our method maintains functional regions for a room, which gather semantically related objects in the same region.
In addition, in the hierarchical scene representation, we model the spatial relationships between the objects in the same region.
Thus, the hierarchical scene representation provides a reasonable room structure at the semantic level, which makes the result has clear functional division and harmonious object placements.
In contrast, HoloDeck's result does not have a clear division of regions.
In HoloDeck's result of Figure~\ref{fig:hierarchy-effective}, the placements of objects in A, B, and C lack interaction with surrounding objects.
Besides, in D and E, the sofas face the walls and also lack interaction with other objects.

\subsection{User Study}
\begin{table}[htbp]
\centering
\SetTblrInner{rowsep=3pt}
\caption{
User study results comparing five methods across five evaluation aspects. *VLM Avg. is derived from the Table 3 of the main body.
}
\vspace{-2mm}
\begin{tblr}{
colspec={lcccccc|c},
colsep={3pt},
row{1} = {font={\bfseries \footnotesize}},
rows = {font=\footnotesize},
cells={halign=c,valign=m},
cell{1}{1,2} = {c},
hline{2,6}={1-9}{.5pt, solid},
hline{1,7}={1-9}{1pt, solid},
}
Method & PR$\uparrow$ & SR$\uparrow$ & LR$\uparrow$ & IA$\uparrow$ & AC$\uparrow$ & {Human \\ Avg.$\uparrow$} & {*VLM \\ Avg.$\uparrow$} \\
Holodeck~\cite{HoloDeck} & 61.7 & 61.5 & 60.6 & 61.4 & 60.1 & 61.1 & 62.3 \\
LayoutVLM~\cite{LayoutVLM} & 47.4 & 48.6 & 49.7 & 49.8 & 49.2 & 48.9 & 61.2\\
Deng \etal~\cite{Global-local} & \underline{63.7} & \underline{61.7} & \underline{62.7} & \underline{62.2} & \underline{64.9} & \underline{63.0} & \underline{70.5} \\
DirectLayout~\cite{DirectLayout} & 60.4 & 60.3 & 60.3 & 59.5 & 59.0 & 59.9 & 64.7 \\
Ours & \textbf{66.7} & \textbf{67.8} & \textbf{66.7} & \textbf{67.1} & \textbf{66.7} & \textbf{67.0} & \textbf{80.5} \\
\end{tblr}
\label{tab:user_study}
\end{table}

To assess the perceptual quality of generated scenes, we follow HoloDeck~\cite{HoloDeck} and LayoutVLM~\cite{LayoutVLM} to conduct a human evaluation study. 
We randomly sample a subset with 120 results generated by different methods in our 3DTindo-Bench.
Each of the 120 test samples is independently annotated by six graduate students, who rank five methods (HoloDeck~\cite{HoloDeck}, LayoutVLM~\cite{LayoutVLM}, Deng \etal~\cite{Global-local}, DirectLayout~\cite{DirectLayout} and our method) across five aspects: PR, SR, LR, IA, and AC, which are consistent with our 3DTindo-Bench's evaluation aspects.
For each aspect, annotators are presented with all five scene renderings and asked to sort them from worst to best. 

Each method received a normalized score computed as:
\begin{equation}
  S=100*\frac{r}{5},
\end{equation}
where $r$ is the rank of a method across five methods and $S$ is the normalized human score.
A higher score indicates better alignment with the aspect.
Table~\ref{tab:user_study} reports the human evaluation scores.
Our method outperforms all baselines across every aspect.
Moreover, it shows a strong correlation of the ratings between human evaluation and the VLM evaluation in Table 3 in the main body.
Therefore, our 3DTindo-bench's VLM-based evaluation is capable of simulating human perception.

\begin{figure*}
    \centering
    \includegraphics[width=1\linewidth]{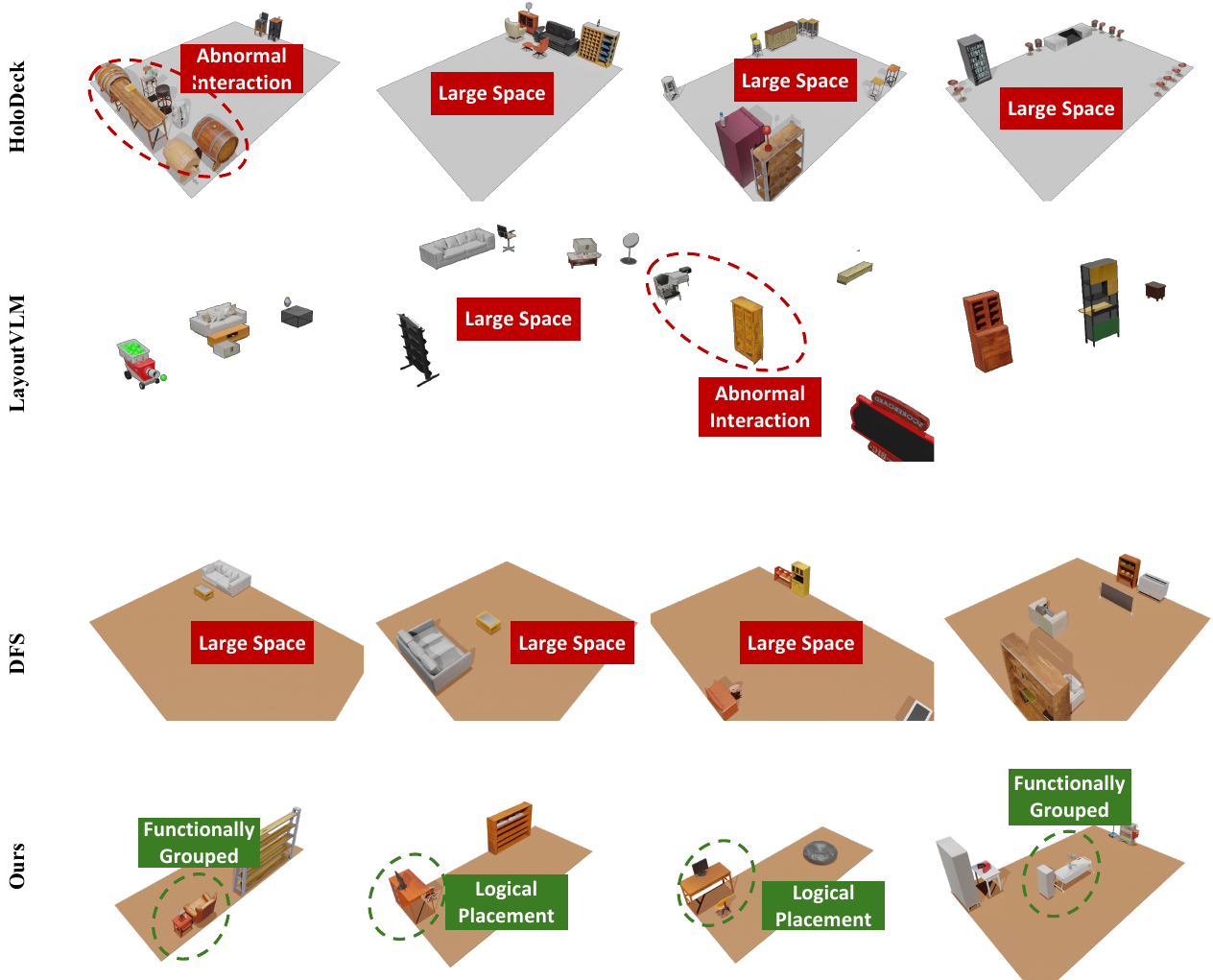}
    \caption{Qualitative results HoloDeck, LayoutVLM, DFS, and our method.}
    \label{fig:vis-1}
\end{figure*}

\begin{figure*}
    \centering
    \includegraphics[width=1\linewidth]{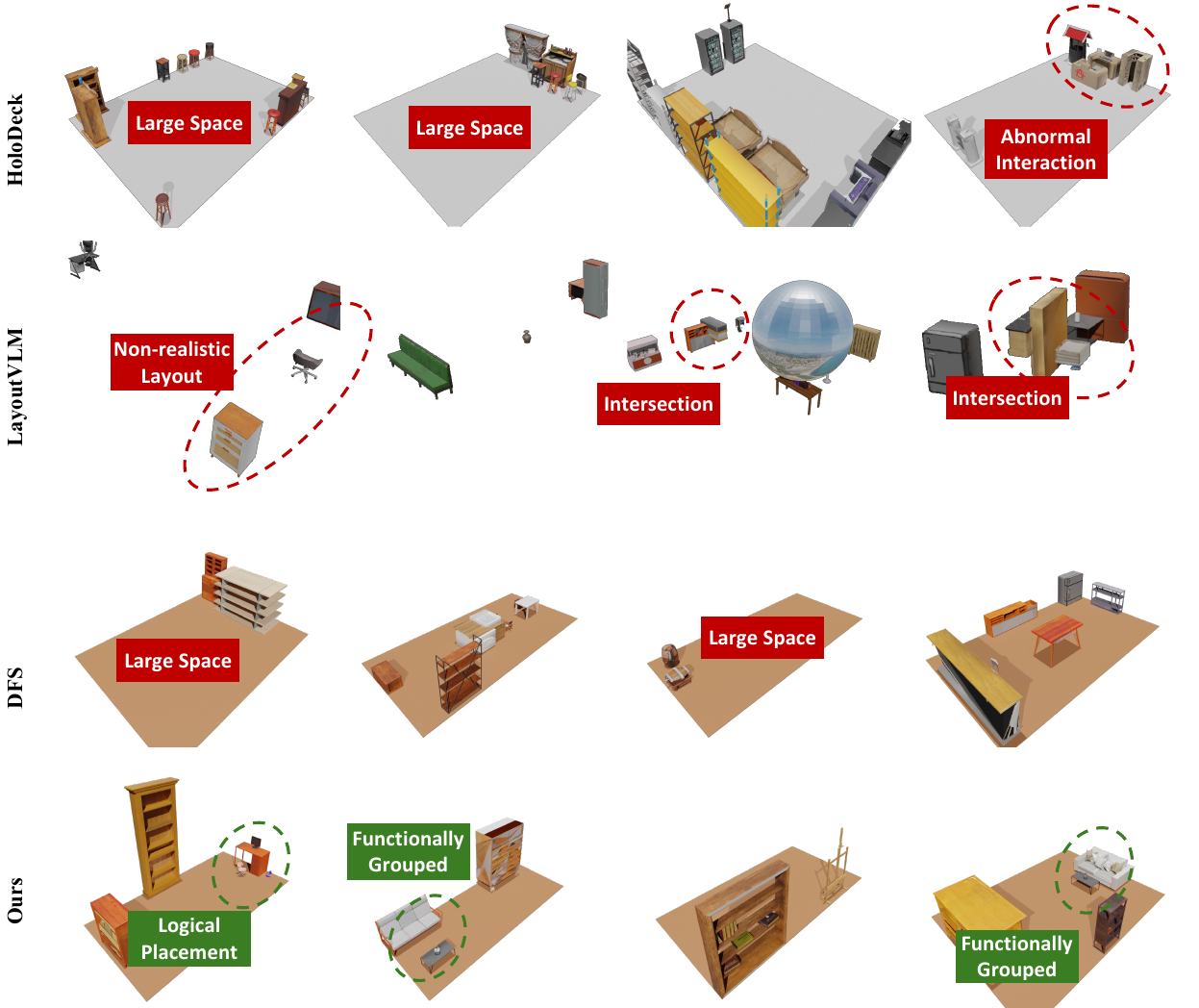}
    \caption{Qualitative results HoloDeck, LayoutVLM, DFS, and our method.}
    \label{fig:vis-2}
\end{figure*}

\noindent\textbf{More qualitative results.}
We visualize more comparison results with HoloDeck~\cite{HoloDeck}, LayoutVLM~\cite{LayoutVLM}, DFS-based tree search~\cite{Global-local}, and MCTS-based tree search (ours) in Figure~\ref{fig:vis-1} and Figure~\ref{fig:vis-2}.

\begin{figure*}[ht]
\centering
\begin{promptbox}[Room-level Node,label={fig:get-room-node}]
You are an expert interior designer tasked with analyzing a room layout. Based on the provided room type, room description, and the description of the room size.

Your job is to estimate the size of the room.

Input:

- Room Description: \{room\_desc\}

You need to estimate a resonable size for the room, based on the description of the size, formatted as (length, width, height) in meters.
\end{promptbox}
\end{figure*}

\begin{figure*}[ht]
\centering
\begin{promptbox}[Region-level Node,label={fig:get-region-node}]
You are an intelligent assistant tasked with categorizing functional areas in a room based on its type, description, and size. Your goal is to carefully analyze the provided information and determine which functional areas from the list are suitable for the given room. Follow these steps to ensure accurate results:

1. **Evaluate Room Description**: Read the detailed description of the room to understand its features, layout, and size.

2. **Determine Room Size**: Assess the size of the room (e.g., small, medium, large) based on the description.

3. **Select Functional Areas**: From the provided functional areas, choose one or more that best fit the room type and size. Ensure that:

   - For small rooms, prioritize practicality and consider limiting to one or two functional areas.

   - For medium and large rooms, you can select multiple functional areas if they can be accommodated comfortably.

4. **Describe Spatial Relationships**: For each selected functional area, explain its position relative to others (e.g., to the left, behind, adjacent).

Here are some example functional areas. You can choose from them but not limited to them:

- Rest Area

- Dining Area

- Storage Area

- Work Area

- Cooking Area

- Bathing Area

- etc.

The relationships you can choose from are:

- left of

- right of

- behind

- in front of

Input:
1. Room Description: \{room\_desc\}
\end{promptbox}
\end{figure*}

\begin{figure*}[ht]
\centering
\begin{promptbox}[Layout of regions,label={fig:get-region-layout}]
You are an expert interior designer tasked with analyzing a room layout. Based on the provided room type, size, description, existing functional areas and corresponding descriptions, your job is to determine the layout of the functional regions in the room.

Input:

- Room Description: \{room\_desc\}

- Functional Regions: \{areas[0]\}

- Main Region: \{anchor\_region\}

You need to determine the layout of the functional regions. The following options describe the relation of the regions to the room. You need to specify distinct a index (an integer) for each region:
\{num\_region\_2\_relation[areas[2]]\}

Then, you need to determine a reasonalbe length of each region in meters. 

You should consider the length of the entire room.

You should make the length of the main region (i.e. \{anchor\_region\}) not shorter than the others.

You should not make the regions too short or too long.

Make sure sum of the lengths of each region is equal to the length of the room. (i.e. l\_1 + l\_2 + ... + l\_3 = L\_total)

Here is an example:
---
Input:
- Room Type: living room
- Room Description: a warm and inviting living room
- Room Size: 
  - large: length = X meters
- Functional Regions:
  - Rest Area: The main area for relaxation, watching TV, chatting, or unwinding
  - Dining Area: For daily meals or entertaining guests
  - Storage Area: Storing everyday items, books, electronic devices, or decorations, keeping the living room organized

Output:
  - Rest Area: right side, length = x\_1 (meters)
  - Dining Area: center, length = x\_2 (meters)
  - Storage Area: left side, length  = x\_3 (meters)
---

\end{promptbox}
\end{figure*}

\begin{figure*}[ht]
\centering
\begin{promptbox}[Floor objects, label={fig:get-floor-objects}]
Task:

You are an expert in interior design, and your task is to add furniture for a specific function area within a defined room type.

Input:

Functional Area: \{area['name']\}

Functional Area Description: \{area['desc']\}

Functional Area Size: \{size[0]\} meters length x \{size[1]\} meters width

Please follow these guidelines:

1. Do not provide rug/mat, windows, doors, curtains, floor, and ceiling objects which have been installed for each room.

2. You need to provide a description for each furniture. The description should include the type, function and appearance.

3. You need to estimate a reasonable size (in meters) for each furniture. For the size of each furniture, you should consider the size of the functional area. The size of the furniture should be reasonable and should not be too large.

4. Based on the function of the input furniture, determine the head of this furniture in the top-down view from the following options:

  - longer: means the furniture's head (frontal side) is commonly the longer side in the top-down view rather than the shorter side in top-down view for functional usage. For example: 

    - office table: In the top-down view, the longer side of an office table is used to provide working space for users. Thus, the longer side of an office table is commonly regarded as the frontal side.

  - shorter: means the furniture's head (frontal side) is commonly the shorter side in the top-down view rather than the longer side in top-down view for functional usage. For example: 

    - bed: In the top-down view, the shorter side of a bed is used to place pillows and support head and neck. Thus, the shorter side is typically regarded as the head (frontal side) of a bed.

    - bathtub: In the top-down view, the shorter side of a bathtub is used to place the head of users. Thus, the shorter side of a bathtub is typically regarded as the head (frontal side).

  (If the furniture is not in the examples, you can choose 'longer' for this task.)

5. You must provide main furniture for the area. The main furniture is the one that is typically used in the area and it can present the most important features of the area. For example:

  - a bed for the resting area of a bedroom

6. Do not use plural words in your response. Use singular form for all words (e.g., use "chair" instead of "chairs").

7. The furniture must typically in this specific functional region of the room.

8. The furniture must typically be placed directly on the floor or ground. The object that is usually placed on other furniture is not allowed.

  - item that is typically placed on the floor. This item should not be supported by other furniture. Examples include:

    - sofa

    - bed

  - item that is placed on other furniture. Examples include:

    - television (on television stand)

    - desk lamp (on desk)

9. You must consider the size of the area. The number of furniture you generated should not be too many that the area cannot fit comfortably.

10. If the area has enough space, you should provide as much furniture as possible.

\end{promptbox}
\end{figure*}

\begin{figure*}[ht]
\centering
\begin{promptbox}[Anchor object determination, label={fig:get-anchor}]

You are tasked with classifying objects in a given room based on their role as either Anchor object or Other objects. An Anchor object is defined as a large item that represents the main function of a specified functional area. Other objects do not serve this primary role. Only 1 object can be classified as Anchor object.

Input:

- Room Description: \{room\_desc\}

- Functional Area: \{area['name']\}

- Functional Area Description: \{area['description']\}

- Input Objects: \{objects[0]\}

Please follow these steps:

1. Read the provided room type and description.

2. Understand the functional area and its description.

3. Analyze the list of objects in the functional area.

4. You MUST select 1 Anchor object from the given objects list based on its size and relevance to the main function of the area. Even if none of the objects perfectly fits the role of an Anchor object, you must still select the one that best serves this purpose.

5. Create a JSON output that categorizes each object accordingly.

\end{promptbox}
\end{figure*}

\begin{figure*}[ht]
\centering
\begin{promptbox}[Anchor object placement rule, label={fig:get-anchor-placement}]
You are an expert interior designer tasked with analyzing a room layout. Based on the provided room type, description, a functional region and corresponding description, and an anchor furniture which represents the main function of the input functional region. Your job is to determine the size and placement policy for the input anchor furniture.

Input:

- Functional Region: \{area\_name\}

- Functional Region Description: \{area\_desc\}

- Functional Region Size: length = \{area\_dimension[0]\} meters, width = \{area\_dimension[1]\} meters

- Anchor Furniture: \{obj\_name\}

- Anchor Furniture Description: \{obj\_desc\}

You should decide on the placement rule for this anchor furniture. You can use only the following anchor rules:

(1) "place\_center" which places the anchor furniture at the center of the room.

(2) "place\_wall" which places the anchor with its back against a segment of the wall.

(3) "place\_corner" which places the anchor at a corner.
\end{promptbox}
\end{figure*}

\begin{figure*}[ht]
\centering
\begin{promptbox}[Affiliated object placement rule, label={fig:get-affiliated-placement}]
You are an expert interior designer tasked with analyzing a room layout. Based on the provided room type, description, a functional region and corresponding description, an anchor furniture which represents the main function of the input functional region, and some affiliated furnitures which presents the secondary function of the region. Your job is to determine the size and placement policy for the affiliated furnitures.

Input:

- Functional Region: \{area\_name\}

- Functional Region Description: \{area\_desc\}

- Anchor Furniture: \{anchor\_name\}

- Anchor Furniture Description: \{anchor\_desc\}

- Affiliated Furniture: \{obj\_name\}

- Affiliated Furniture Description: \{obj\_desc\}

[Basic Requirements]

Follow these steps carefully to ensure the task is completed with clarity and accuracy:

1. **Understand the Context**: Based on the input, first assess the situation and explain how you are approaching the task.

2. **Break Down the Problem**: Identify key components of the task and explain the reasoning behind each step.

3. **Step-by-step Execution**: For each step, describe what you're doing, and why, providing reasoning behind decisions made. Ensure that intermediate outputs are included at each stage.

Step 1: You need to determine the placement policy for each affiliated furniture. A placement policy means where should the affiated furniture be placed relative to the anchor furniture (\{anchor\_name\}). You must consider the common sences of the indoor layout design. You should consider the functionality of the furniture. The answer must be chosen from the following options:

(1) "place\_front" which places the furniture in front of the anchor funiture (\{anchor\_name\}). For example:

  - TV stand in front of a sofa.

(2) "place\_beside" which places the furniture beside the anchor funiture (\{anchor\_name\}). For example:

  - Nightstand beside a bed.

(3) "place\_around" which places the furniture around the anchor funiture (\{anchor\_name\}). For example:

  - chair around a dining table.

Step 2: You need to determine if the affiliated furniture and the anchor furniture are highly functionally grouped together. That means users usually put them together as well as use them together.

For example:

  - Bed and Nightstand: The bed and nightstand are functionally grouped together for users to sleep in the bed and reach the nightstand easily.

Most pair of furnitures are not functionally grouped together. If the input furnitures are not present in the "functionally grouped" examples, you should choose "no".

Step 3: You need to determine if the affiliated furniture should be placed close to the wall or in the middle of the room. For example:

  - Close to the wall:
    - Sofa

  - Not close to the wall:

    - armchair

Step 4: You need to describe the distance of the two furnitures when putting them together. You should choose from the following options:

- "adjacent\_to": means the two furnitures are placed close to (adjacent to) each other. For example:

  - nightstand and bed. The nightstand is usually placed adjacent to the bed for users to relax in the bed and easily reach the nightstand.

- "near": means the two furnitures are placed near each other with a space between them. They are not close to (not adjacent to) each other. For example:

  - sofa and coffee table. The coffee table is usually placed near the sofa for users to relax on the sofa and read a book or drink coffee on the coffee table. It needs a space between them to walk around.

- "far": means the two furnitures are placed far from each other. For example:

  - sofa and TV stand. The TV stand is usually placed far from the sofa for comfortable viewing distance.

Step 5: You need to determine the alignment principle of the two furnitures. You need to choose from the following options:

- side alignment: the two furnitures are aligned by their sides. For example:

  - bed and nightstand: the side of the bed is aligned with the side of the nightstand for users to easily access the nightstand when they are in bed.

- center alignment: the two furnitures are aligned by their center. For example:

  - sofa and coffee table: the center of the sofa is aligned with the center of the coffee table for users to easily access the coffee table when they are sitting on the sofa.

- around: the affliated furniture is around the anchor furniture. For example:

  - dining table and chairs: the chairs are around the dining table for users to sit around the table.

\end{promptbox}
\end{figure*}

\begin{figure*}[ht]
\centering
\begin{promptbox}[Side determination, label={fig:side-determination}]
[Role]

You are a professional indoor designer. You know how to place furnitures in a room.

[Task]

Your task is to determine which side of the anchor furniture can accommodate the new furniture.

[Input]

Anchor Furniture Name: \{anchor\_name\}, which is filled with red color and labled with its name.

Input Image: This image presents a floor of a room. The floor of the room is divided into a grid of cells. In this room, there is some furnitures. The furnitures are filled by red and labled with their names. In the free space, the cells are filled by emojis. Besides, the cells in the same {axis} share the same emoji. Around the floor, there are walls and boundaries. The walls are filled with the `wall` (which looks like a brick) emoji and the boundaries are filled with the `boundary` (which looks like a white circle) emoji. Both walls and boundaries cannot be occupied by furnitures. The walls and boundaries are different. The walls are the walls of a room and cannot be passed through. The boundaries are the boundaries of a region in the room and can be passed through.

[Basic Requirements]

Follow these steps carefully to ensure the task is completed with clarity and accuracy:

1. **Understand the Context**: Based on the input, first assess the situation and explain how you are approaching the task.

2. **Break Down the Problem**: Identify key components of the task and explain the reasoning behind each step.

3. **Step-by-step Execution**: For each step, describe what you're doing, and why, providing reasoning behind decisions made. Ensure that intermediate outputs are included at each stage.

4. **JSON Outputs**: After you output the intermediate results. You need to output a JSON.

[Task Requirement]

1. I want to introduce a new furniture, "\{target\_name\}", size = \{target\_size\} \{axis\}s, into the room in the input image.

2. Please choose the side of the anchor furniture, "\{anchor\_name\}" that is filled with red color and labled with its name, that can accommodate the new furniture. The side you choose should have enough \{axis\}s for the new furniture. The new furniture requires \{target\_size\} \{axis\}s.

3. You are required to only choose from the following options and do not output other options:

  - \{direction\_options\}

  (if both of them are suitable, you can choose either one.)

4. The cells that are filled with emojis except 'brick' and 'white circle' are available for the new furniture. The cells that are filled with 'brick', 'white circle' and are not filled with any emoji are not available for the new furniture.

5. You should first output the reasoning for the decision. Specifically, you should describe the input image in detail, including the layout, emojis, walls, boundaries, etc. Describe the emoji at the top, bottom, left, right of the anchor furniture. Describe the real color of the `bed`  

\end{promptbox}
\end{figure*}

\begin{figure*}[ht]
\centering
\begin{promptbox}[Row/Column determination, label={fig:row-column-determination}]
[Role]
You are a professional indoor designer. You know how to place furnitures in a room.

[Task]
Your task is determine the {axis}s that the furniture should be placed.

[Input]
Anchor Furniture Name: \{anchor\_name\}, which is filled with red color and labled with its name.
Input Image: This image presents a floor of a room. The floor of the room is divided into a grid of cells. In this room, there is some furnitures. The furnitures are filled by red and labled with their names. In the free space, the cells are filled by emojis. Besides, the cells in the same \{axis\} share the same emoji. Around the floor, there are walls and boundaries. The walls are filled with the `wall` (which looks like a brick) emoji and the boundaries are filled with the `boundary` (which looks like a white circle) emoji. Both walls and boundaries cannot be occupied by furnitures. The walls and boundaries are different. The walls are the walls of a room and cannot be passed through. The boundaries are the boundaries of a region in the room and can be passed through.
Distance Requirement: \{distance\_desc\}. Explain:

- if "adjacent\_to": the new furniture should be placed close to (adjacent to) each anchor furniture.

- if "near": the new furniture should be placed near the anchor furniture but not close to (adjacent to) the anchor furniture.

- if "far": the new furniture should be  placed far from the anchor furniture.

[Basic Requirements]

Follow these steps carefully to ensure the task is completed with clarity and accuracy:

1. **Understand the Context**: Based on the input, first assess the situation and explain how you are approaching the task.

2. **Break Down the Problem**: Identify key components of the task and explain the reasoning behind each step.

3. **Step-by-step Execution**: For each step, describe what you're doing, and why, providing reasoning behind decisions made. Ensure that intermediate outputs are included at each stage.

4. **JSON Outputs**: After you output the intermediate results. You need to output a JSON.

[Task Requirement]

1. I want to introduce a new furniture, "\{target\_name\}", size = \{target\_size\} \{axis\}s, into the room in the input image.

2. The new furniture should be placed \{direction\_desc\} the anchor furniture, "\{anchor\_name\}". You need to pay attention to the distance requirement.

3. The new furniture should be placed in the free space and should not overlap with any other furnitures.

4. Please describe the emojis that the furniture should be placed. Pay attention:

  - You can only describe the emojis with their names from the following list. The length of the list is \{len(emoji\_used)\}: \{emoji\_used\}

  - The length of emojis you provided must equals \{target\_size\}, standing for the number of \{axis\}s the furniture need to be occupied.

  - The emojis you provided must be distinct.

  - The emojis you provided must be adjacent to each other.

\{close\_to\_the\_wall\}
5. You should first output the reasoning for the decision.

6. If no suitable position is found, the answer should be None.
\end{promptbox}
\end{figure*}

\end{document}